%% file: main_arxiv.tex
\documentclass[runningheads]{llncs}

\usepackage{eccv}

\usepackage{eccvabbrv}

\usepackage{graphicx}
\usepackage{booktabs}
\usepackage{multirow}
\usepackage[accsupp]{axessibility}  

\usepackage[pagebackref,breaklinks,colorlinks]{hyperref}

\usepackage{orcidlink}
\usepackage{todonotes}
\usepackage{graphicx}
\usepackage{cite}
\usepackage{wrapfig}
\usepackage{tabularx}
\usepackage{bm}
\usepackage[perpage]{footmisc}

\begin{document}

\title{Topo4D: Topology-Preserving Gaussian Splatting for High-Fidelity 4D Head Capture}

\titlerunning{Topo4D}

\author{Xuanchen Li \inst{1} \and
Yuhao Cheng\inst{1} \and
Xingyu Ren\inst{1}
\and
Haozhe Jia\inst{2}
\and
Di Xu\inst{2}
\and
\\
Wenhan Zhu\inst{3}
\and
Yichao Yan\inst{1\dag}
}

\institute{MoE Key Lab of Artificial Intelligence, AI Institute, Shanghai Jiao Tong University \email{ \{lixc6486,chengyuhao,rxy\_sjtu,yanyichao\}@sjtu.edu.cn}
\and
Huawei Cloud Computing Technologies Co., Ltd
\email{\{jiahaozhe1,xudi21\}@huawei.com}
\and
Xueshen AI
\email{whzhu@foxmail.com}
}

\authorrunning{Xuanchen Li et al.}


\maketitle

\input{sections/abs} 
\footnotetext[0]{\textsuperscript{$\dag$}Corresponding author}

   
\input{sections/intro}
\input{sections/relatedworks}
\input{sections/method_v3}

\input{sections/exp}
\input{sections/conclusion}

%
%
\bibliographystyle{splncs04}
\bibliography{main}

\clearpage

\appendix
\renewcommand{\theequation}{\Alph{equation}}
\renewcommand{\thefigure}{\Alph{figure}}
\renewcommand{\thetable}{\Alph{table}}
\setcounter{equation}{0}
\setcounter{figure}{0}
\setcounter{table}{0}

\input{supp}

\end{document}

%% file: sections/abs.tex
\begin{abstract}
Recent significant advances in high-quality face reconstruction have been made, but challenges remain in 4D face asset reconstruction.
4D head capture aims to generate dynamic topological meshes and corresponding texture maps from videos, which is widely utilized in movies and games for its ability to simulate facial muscle movements and recover dynamic textures in pore-squeezing.
The industry often adopts a method involving multi-view stereo and non-rigid alignment. However, this approach is prone to errors and heavily relies on time-consuming manual processing by artists.
To simplify this process, we propose Topo4D, a novel framework for automatic geometry and texture generation that optimizes densely aligned 4D heads and 8K texture maps directly from calibrated multi-view time-series images.
Specifically, we first represent the time-series faces as a set of dynamic 3D Gaussians with fixed topology in which the Gaussian centers are bound to the mesh vertices.
Afterward, we optimize geometry and texture frame-by-frame alternatively for dynamic head capture
while maintaining temporal topology stability.
Finally, we can extract dynamic facial meshes in regular wiring arrangement and high-fidelity textures with pore-level details from the learned Gaussians.
Extensive experiments show that our method achieves superior results than the current SOTA face reconstruction methods in the quality of both meshes and textures. Project page: \url{https://xuanchenli.github.io/Topo4D/}.

\keywords{4D Face Modeling \and High Resolution Texture Generation 
}
\end{abstract}

%% file: sections/intro.tex
\section{Introduction}

\input{Figures/teaser}

4D head capture requires obtaining temporal-continuous topological facial assets, including facial geometries and textures. It has been widely used in entertainment media, such as games, movies, and interactive AR/VR, to create dynamic faces with realistic and immersive quality. The main challenges of 4D head capture are: \textbf{1)} representing faces with a fixed \textbf{topology} and a regular UV, and \textbf{2)} maintaining \textbf{temporal stability} among different frames.

To produce captivating and lively 4D facial assets, the industrial pipeline typically employs professional equipment, \eg, Light Stage~\cite{debevec2012light}, to capture high-quality multi-view videos. Then multi-view stereo (MVS)~\cite{goesele2006multi,ma2007rapid} is used to compute the facial scan of each frame, followed by a non-rigid registration~\cite{ICP} process to superimpose the topologically aligned faces onto the scans. To achieve temporal consistency and obtain usable assets, this process requires marking on the subject's face and a manual post-process by artists. To eliminate the need for manual operations, some methods~\cite{beeler2011high,bradley2010high,fyffe2017multi, yang2020facescape} employ optical flow or other techniques as supervision to automatically warp template models at the expense of processing time. Additionally, they necessitate careful parameter tuning for different subjects to achieve optimal results.
Therefore, there is an urgent demand to develop more automated workflows to accelerate the 4D asset reconstruction.

To achieve automatic and efficient facial reconstruction, researchers have developed deep-learning models for \textbf{topological} facial asset generation frame-by-frame, which can be divided into two categories.
The first line of work~\cite{Ganfit,DECA, aldrian2012inverse, bas2017fitting, ploumpis2020towards,FLAME, DFNRMVS, MVFNet,hifi3dface2021tencentailab, thies2016face2face, wood20223d, 3DDFA}
utilizes parametric models to fit the facial images for mesh creation, where parametric texture or inverse rendering is employed for corresponding texture generation. These methods are highly efficient, but due to their limited expressive abilities, they struggle to produce high-quality textures with diverse identities and complex expressions. 
Another line of works~\cite{TOFU,TEMPEH,REFA,DFNRMVS,MVFR,MVFNet} proposes to directly regress face models from multi-view images, where a large amount of expensive pre-processed 3D data are utilized for training. 
Typically, these methods can capture consistent features across multiple views to predict accurate geometry, while several recent works~\cite{Lattas20,REFA} successfully generate high-resolution textures with a super-resolution module.
Nevertheless, super-resolution modules often introduce artifacts and fail to faithfully recover the details of faces. 
Furthermore, these two kinds of aforementioned methods encounter the same predicament in that they are designed to reconstruct each frame individually without directly applying frame continuity, thereby struggling to maintain the temporal coherence between frames.

The recent progress in 3D Gaussian Splatting~(3DGS)~\cite{3DGS} and its advancements in 4D scene representation~\cite{DynamicGaussian,yang2023deformable,wu20234d,yang2023real,liang2023gaufre} bring us inspiration. These methods of high-fidelity 4D scene representation fully take into account the continuity among frames and achieve \textbf{temporal-consistent} reconstruction. Moreover, thanks to its advanced rendering pipeline, it can achieve ultra-high resolution texture learning and fast perspective rendering in a memory-efficient manner. Considering their success in 4D scene representation, it would be highly desirable if we could extract high-fidelity facial meshes and textures in the pre-defined topology from these 3DGS-inspired representations. However, it is non-trivial since the Gaussians in these representations are random and uncontrollable, making it difficult to perfectly register geometries with a fixed topology. 

To overcome these challenges, we propose a novel optimization framework, \textbf{Topo4D}, to obtain high-fidelity 4D meshes and temporal-stable textures.
\textbf{1)}~To maintain the quality of photo-realistic rendering in 3DGS and also extract meshes and textures with \textbf{fixed topology}, we first explicitly link 3D Gaussians to the pre-defined topological facial geometry, named Gaussian Mesh. Hence, it strikes a balance between high expressive power and geometric structure fixedness to help learn topology-constrained meshes and photo-realistic textures. 
\textbf{2)}~Considering that the topology may be destroyed during dynamic optimization, we design physical and geometric prior constraint terms on topological relationships during the optimization process to ensure \textbf{temporal topological stability} and regular mesh arrangement. Moreover, inheriting the advantages of 3DGS, the optimization process is computationally fast and memory-efficient.
\textbf{3)} Once optimized, we align the Gaussian surface with the rendered surface using geometric normal prior to extracting high-quality meshes. In addition, we design the UV densification module to learn ultra-high-resolution textures with pore-level details by inversely mapping the Gaussian colors into UV space.

Our approach's effectiveness has been validated through extensive experiments. To our knowledge, Topo4D is the first method to implement the generation of high-fidelity 4D facial models with native 8K texture mapping, demonstrating the potential of Gaussian for dynamic face reconstruction and ultra-high-resolution tasks.
In summary, our contributions include:
\begin{itemize}
    \item We propose a novel optimization framework, Topo4D, for the reconstruction of high-quality 4D heads and photo-realistic textures with pore-level details from multi-view videos.
    \item We propose the Gaussian Mesh with UV densification to better represent facial models in the pre-defined topology and fixed UV.
    \item We design the alternative geometry and texture optimization process to ensure temporal topology stability and regular mesh arrangement during the optimization process.
    
\end{itemize}

%% file: Figures/teaser.tex
\begin{figure*}[t]
    \centering
    \includegraphics[width=\linewidth]{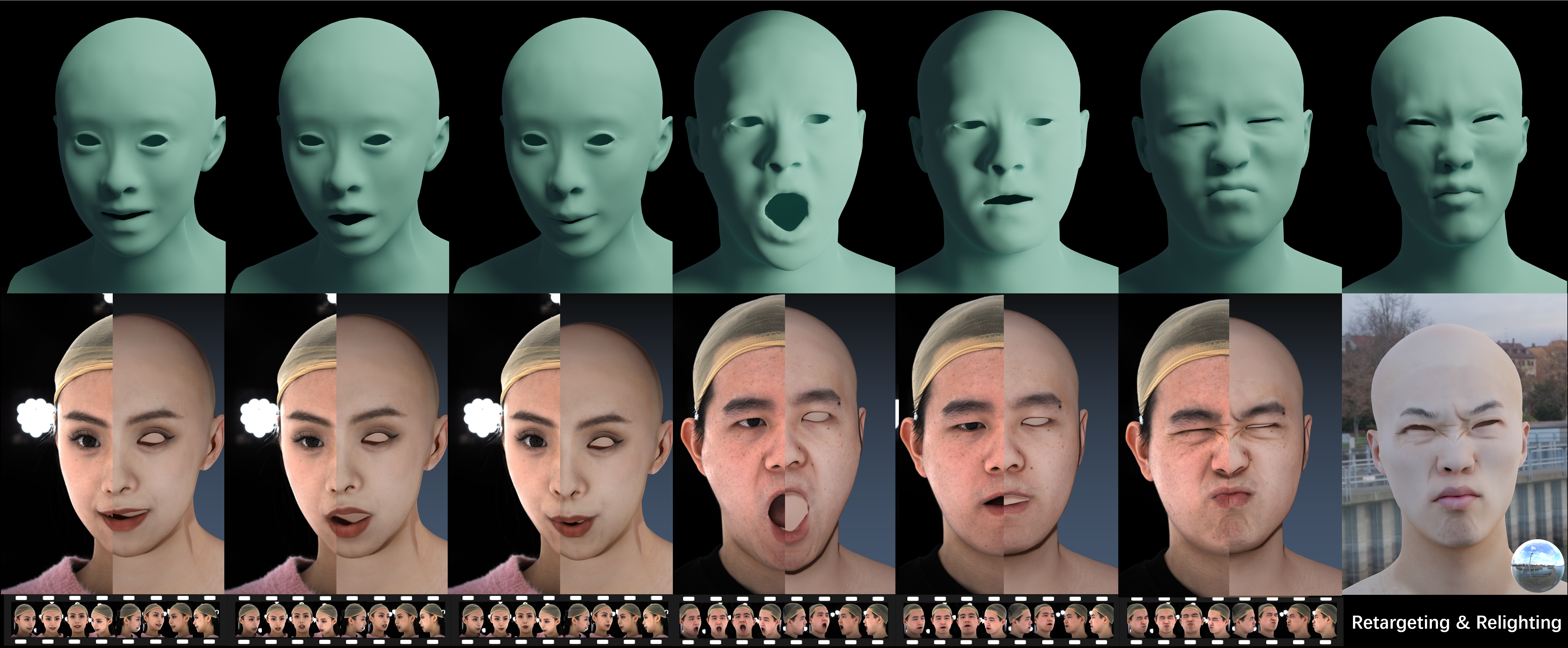}
    \captionof{figure}{Example results of our Topo4D
    . Our method can produce temporal-consistent topological head meshes with high-fidelity 8K textures from calibrated multi-view videos. Captured 4D models can be applied to retargeting and relighting applications.}
    \label{fig:teaser}
\end{figure*}

%% file: sections/relatedworks.tex
\section{Related Works}
\subsection{Registered Facial Model Acquisition}
Acquiring high-fidelity facial models in pre-defined topologic structures has been a long-standing research challenge. Beginning with 3D Morphable Models~\cite{3DMM}, many methods~\cite{Ganfit,DECA, aldrian2012inverse, bas2017fitting, ploumpis2020towards,FLAME, DFNRMVS, MVFNet,hifi3dface2021tencentailab, thies2016face2face, wood20223d} employ parametric models to achieve face reconstruction with single- or multi-view images. However, such methods struggle to faithfully reconstruct faces. 
Non-rigid ICP~\cite{ICP, booth20163d, cao2013facewarehouse, ji2021light} methods deform canonical models to fit scans reconstructed by MVS methods with high quality.
However, naively extending these methods to 4D videos will result in temporal instability, leading to texture drift.
To maintain temporal stability, high-precision models are re-topology frame-by-frame in existing CG pipelines using professional software, \eg, Wrap4D~\cite{Wrap4D}, which demands extensive time and expertise from experienced artists.
To solve this, 
some methods~\cite{beeler2011high,bradley2010high,fyffe2017multi} utilize optical flow or other techniques as supervision to deform the template models. However, numerous hyper-parameters need to be carefully tuned in these methods, making it difficult to generalize to different identities, and they are computationally slow.
Another category of methods~\cite{TOFU,TEMPEH,REFA,HRN} directly regresses models from images. while these methods are constrained by a large number of training data and face difficulties in extending to other capture systems. Besides, out-of-domain expressions may be limited by insufficient training data.
In this paper, we propose a novel approach for acquiring registered facial models with the quality typically achieved by artists manually, but in significantly less time.

\subsection{Facial UV-Texture Recovery}
Traditional CG pipelines predominantly use inverse rendering on reconstructed meshes to acquire textures from images at a computationally slow speed.
To accelerate the process, many methods~\cite{deng2018uv,Lattas20,ren2023facial,UnsupTex,HRN,zhang2022video,lattas2023fitme,hifi3dface2021tencentailab} directly extract features from images to generate textures, where the quality is limited by resolution. 
Despite employing super-resolution networks~\cite{Lattas20,REFA}, they may still arise artifacts on the texture, and cannot accurately replicate pore-level details.
Moreover, the aforementioned approaches are primarily designed for static tasks, and thus may not ensure the temporal stability of textures.
Notably, Zhang et al.~\cite{zhang2022video} can achieve video-level texture generation. However, it can only produce wrinkle maps that can be composited with natural high-resolution textures to represent varied expressions, rather than directly generating textures with high-frequency details, limiting its availability.
Compared to these approaches, our method ensures temporal topological consistency in texture recovery and can directly generate textures in native 8K resolution with pore-level details.

\subsection{Scene Representation}
Neural Radiance Fields~(NeRF)~\cite{mildenhall2021nerf} has garnered significant attention for its remarkable capability to faithfully preserve both geometric and texture details of objects.  
Subsequent advancements in training speed~\cite{muller2022instant,fridovich2022plenoxels,chen2022tensorf}, inference speed~\cite{lin2022efficient,chen2023mobilenerf,lombardi2021mixture}, geometric quality~\cite{wang2021neus,yariv2023bakedsdf,meng2023neat}, rendering quality~\cite{barron2021mip,barron2022mip,jiang2023alignerf}, and dynamic scene representation~\cite{gao2021dynamic,li2022neural,fridovich2023k, dialoguenerf} have considerably enhanced the applicability of NeRF. 
Furthermore, 3D Gaussian splatting~(3DGS)~\cite{3DGS} has achieved SOTA results in scene representation due to its high-fidelity rendering, efficient training, and inference speeds, as well as memory efficiency.
This technique has been further extended to 4D scene reconstruction~\cite{DynamicGaussian,yang2023deformable,wu20234d,yang2023real,liang2023gaufre}, with notable applications in representing dynamic heads~\cite{GaussianAvatar,xu2023gaussian,chen2023monogaussianavatar} and bodies~\cite{li2023animatable,zielonka2023drivable,qian20233dgs}. 
However, these methods typically utilize continuous neural networks or random Gaussians to represent objects, posing challenges in extracting meshes with fixed topological structures, thus limiting their integration with existing industrial processes. 
To address these issues, we propose a meticulously improved 3DGS framework that can extract dynamic high-quality meshes and photo-realistic textures in constrained topology and UV from multi-view videos. Additionally, our method can be directly applied to current computer graphic industrial processes.

%% file: sections/method_v3.tex
\section{Methods}
Our method aims to achieve temporally stable head mesh reconstruction and texture recovery from calibrated multi-view videos. Specifically, given sets of multi-view image sequences $\{ \mathbf{I}^j_i \in \mathbb{R}^{h \times w \times 3} | 0 \leq i \leq F-1\}_{j=1}^K$ in the resolution of $h \times w$, encompassing $F$ frames captured from $K$ different viewpoints, all with known camera calibrations, our method can extract head meshes 
$\{\mathbf{S}_i:=(V^i, T)|V^i\in\mathbb{R}^{n_v\times3} \}_{i=0}^{F-1}$ in the pre-defined fixed topology $T$ together with texture maps $\{\mathbf{M}_i\in\mathbb{R}^{8192\times8192\times3} \}_{i=0}^{F-1}$, where $n_v$ represents the number of vertices.

To begin with, we give a brief review of 3D Gaussian Splatting~\cite{3DGS}~(Sec.~\ref{sec:Preliminary}). Our method first builds a \textit{Gaussian Mesh} by initializing a topology-integrated Gaussian set based on the facial priors in the first frame~(Sec.~\ref{sec:Gaussian Mesh}). Then, for each subsequent frame, we alternatively perform geometry optimization and texture optimization, to learn dynamic high-fidelity geometries and textures~(Sec.~\ref{sec:GTO}).
Finally, we introduce how to extract geometries from \textit{Gaussian Mesh} and recover ultra-high-resolution textures~(Sec.~\ref{sec:GTE}). The full pipeline is illustrated in Fig.~\ref{fig:pipeline}.

\input{Figures/pipeline}

\subsection{Preliminary}
\label{sec:Preliminary}
3D Gaussian Splatting (3DGS)~\cite{3DGS} is proposed as a competitive solution for photo-realistic rendering. 
Different from other implicit methods, 3DGS explicitly maintains a set of Gaussian distributions to model a scene. 
In 3DGS, each ellipsoidal Gaussian features a learnable color component $\bm{c}$ and an opacity component $\bm{\sigma}$, and is described by a covariance matrix $\Sigma$ and its mean position $\bm{\mu}$:
\begin{equation}
G(\bm{x})=e^{-\frac{1}{2} (\bm{x} - \bm{\mu})^T\Sigma ^{-1}(\bm{x} - \bm{\mu})},
\end{equation} 
where $\Sigma$ is further decomposed into rotation matrix $R$, parameterized with a quaternion $\bm{q}$, and scaling matrix $S$ :
\begin{equation}
\Sigma=RSS^TR^T.
\end{equation} 

In rendering, the color $\bm{C}$ of a pixel is acquired by sampling and blending all Gaussians that overlap the pixel in depth order: 
\begin{equation}
\bm{C}=\sum_{i=1}\bm{c}_i\alpha_i\prod_{j=1}^{i-1}(1-\alpha_j),
\end{equation} 
where the blending weight $\alpha_i$
is given by evaluating a 2D Gaussian with its covariance multiplied by its opacity.

However, 3DGS is designed for realistic rendering instead of 3D reconstruction. 
Gaussians lack inherent topological relationships, thus unconstrained optimization of their attributes may only yield irregular geometry. As a result, directly extracting topologically sound meshes becomes unfeasible.

\subsection{Gaussian Mesh for Topology Integrated Gaussians Initialization}
\label{sec:Gaussian Mesh}

Extracting topologically consistent meshes from Gaussians is challenging due to the lack of geometric constraints during optimization.
To this end, we propose \textbf{Gaussian Mesh}, which uniquely incorporates the topological prior into vanilla Gaussian and will not affect its high-fidelity rendering quality.
We define 4D Gaussian Meshes as $G_i=\{G_{i,j}\}_{j=1}^{n_v}$ for the $i$-th frame, where $G_{i, j}$ covers some learnable parameters, \ie, $\{\bm{\mu}_{i, j} \in \mathbb{R}^{3}, \bm{q}_{i, j} \in \mathbb{R}^{4}, \bm{s}_{i, j} \in \mathbb{R}^{3}, \bm{c}_{i, j} \in \mathbb{R}^{3}, \bm{\sigma}_{i, j} \in \mathbb{R}\}$ for the position, rotation, scaling, color, and opacity separately of the $j$-th vertice in pre-defined topology $T$.

Different from 3DGS~\cite{3DGS} using SFM-generated messy sparse points for initialization, to directly obtain pre-defined topological information, we first initialize Gaussian Mesh with head mesh and texture of the first frame, which is acquired by automatic MVS and ICP algorithms. Specifically, we set the mean positions $\bm{\mu}_0$ of Gaussians to the corresponding 3D coordinates of vertices in topological order. Furthermore, 
to better align Gaussians with the surface, we initialize the orientation $\bm{q}_0$ of each Gaussian with the normal direction of vertices. 
It is worth noting that recent GaussianAvatars~\cite{GaussianAvatar} also rigs Gaussians to face model. It aims at driving Gaussians with parametric models for photo-realistic rendering, where meshes are obtained by optimizing FLAME~\cite{FLAME} parameters with the utilization of landmarks, personal displacements, and other additional supervision.
In contrast, our goal is to extract high-quality topologic meshes and textures from multi-view videos by directly tracking the sequences without preprocessing registration and tracking. 

After initializing the shape-related attributes ($\bm{\mu}_0$ and $\mathbf{q}_0$) of Gaussians, we then optimize their rendering-related attributes ($\bm{s}_0$, $\bm{c}_0$ and $\bm{\sigma}_0$).
Concretely, we initialize the scales $\bm{s}_0$ as half of the minimum distance between each Gaussian and its one-ring neighbors and set opacity $\bm{\sigma}_0$ to 1, to avoid color blending between multiple Gaussians. To faithfully represent the color, we also initialize the color $\bm{c}_0$ of each Gaussian by sampling the corresponding pixel on the texture map according to the UV coordinate.
Finally, to learn more details in dense parts, \eg, eyes and mouth, we optimize Gaussian's rotation $\bm{q}_0$ and scale $\bm{s}_0$ between the first frames and rendered images $\mathbf{I}'_0$ with the same loss function as 3DGS:
\begin{equation}
    \label{loss:image}
    \mathcal{L}_{image} = (1-\lambda_{image} ) \mathcal{L}_{1}(\mathbf{I}_0,\mathbf{I}'_0) + \lambda_{image}\mathcal{L}_{D-SSIM}(\mathbf{I}_0,\mathbf{I}'_0),
\end{equation} 

Also, Gaussians have volume, leading to a certain gap between the center of Gaussians and the true surface. 
Therefore, the thickness of the Gaussian in the normal direction should be as small as possible.
Therefore, we propose a scale loss to encourage the minimum value of every Gaussian's scale close to 0 and penalize Gaussians with a scale exceeding $\lambda_{init}$ times than its initial value $\bm{s}_{init, i}$:
\begin{equation}
    \mathcal{L}_{scale} = \sum_{i\in G}(\Vert \bm{s}_{0, i} \Vert_{-\infty} + max(0, \bm{s}_{0, i} -\lambda_{init} \bm{s}_{init, i})).
\end{equation} 

Overall, the final loss to initialize Gaussian Mesh can be formulated as:
\begin{equation}
    \mathcal{L}_{init} = \mathcal{L}_{image} + \lambda_{scale} \mathcal{L}_{scale}.
\end{equation}

\subsection{Alternative Geometry and Texture Optimization}
\label{sec:GTO}
After initializing Gaussian Mesh $G_0$, we propose an \textbf{Alternative Geometry and Texture Optimization method} to acquire Gaussian Mesh geometry and learn dense texture colors.
Specifically, at frame $t$, we first optimize Gaussians $G_t$ by tracking $G_{t-1}$ under the regularization of topology and physics. Afterward, dense texture color can be learned based on the tracked geometry. We perform such an alternative optimization process once per frame.

\noindent{\textbf{Geometry Optimization.}}\label{sec:geo opt} 
Naively optimizing Gaussian Mesh causes topological confusion. To maintain the topology within Gaussians and regular mesh arrangement, we extend 3DGS~\cite{3DGS} by introducing physical and topological priors.
Specifically, we propose physical loss item $\mathcal{L}_{phy}$ that constrain local rigidity and topological loss $\mathcal{L}_{topo}$ that improve the mesh quality.
Overall, the loss functions for geometry optimization are constructed in three parts: 
\begin{equation}
    \mathcal{L}_{geo} = \mathcal{L}_{image} + \mathcal{L}_{phy} + \mathcal{L}_{topo},
\end{equation} 

\noindent{\textit{Physical Prior Loss.}} Solely utilizing color as supervision to optimize Gaussians is disastrous since low-frequency texture details, \eg, forehead and cheek, lead to point mistracking, thus breaking the facial topology. To better regularize the motion of Gaussians and maintain topologic information, we modify the loss functions in Luiten et al.~\cite{DynamicGaussian} with the constraints by one-ring neighbors as:
\begin{equation}
     \mathcal{L}_{rot}=\frac{1}{2n_e} \sum_{i\in G}\sum_{j\in \mathcal{K}_i} w_{i,j}\Vert\bm{\hat q}_{t, j}\bm{\hat q}_{t-1, j}^{-1}-\bm{\hat q}_{t, i}\bm{\hat q}_{t-1, i}^{-1}\Vert_2,
\end{equation} where $\bm{\hat q}$ is the normalized quaternion, $n_e$ is the number of edges, and $\mathcal{K}_i$ means the one-ring neighbours of $G_{t,i}$. The loss weighing factor $w$ takes into account the edge length of the Gaussian Mesh at the first frame:\begin{equation}
    w_{i, j}=exp(-\lambda_w\Vert\bm{\mu}_{0, j}-\bm{\mu}_{0, i}\Vert_2^2).
\end{equation}

In addition to the rotation similarity calculated between adjacent frames, we find that long-term physical loss is important for maintaining long-term stable dense correspondence:
\begin{equation}
    \mathcal{L}_{iso}=\frac{1}{2n_e} \sum_{i\in G}\sum_{j\in \mathcal{K}_i} w_{i,j}|\Vert\bm{\mu}_{0, j}-\bm{\mu}_{0, i}\Vert_2-\Vert\bm{\mu}_{t, j}-\bm{\mu}_{t, i}\Vert_2|.
\end{equation}

Finally, our physical loss items is the weighted sum of these two loss functions:
\begin{equation}
    \mathcal{L}_{phy} = \lambda_{rot}\mathcal{L}_{rot} + \lambda_{iso}\mathcal{L}_{iso}.
\end{equation}

\noindent{\textit{Topology Prior Loss.}} Physical constraints achieve long-term tracking of the corresponding Gaussians, but they will lead to irregular wiring and unsmooth surfaces. 
To realize regular wiring, we calculate the L2 loss between the position of each vertex and the average position of its neighbors:
\begin{equation}
    \mathcal{L}_{pos}=\frac{1}{n_v}\sum_{i\in G}(\bm{\mu}_{t, i}-\frac{ {\textstyle \sum_{j\in \mathcal{K}_i}\bm{\mu}_{t, j}} }{|\mathcal{K}_i|} )^2.
\end{equation}

To maintain the stable topology and smooth surface, we further apply mesh flattening loss to the angles between adjacent faces in optimization:
\begin{equation}
    \mathcal{L}_{flat} = \sum_{\theta_i \in e_i}(1-cos(\theta_{t, i}-\theta_{0, i})),
\end{equation} 
where $\theta_{t, i}$ is the angle between the faces that have the common edge $e_i$ at frame $t$. 
Overall, our topological prior loss servers are as:
\begin{equation}
    \mathcal{L}_{topo} = \lambda_{pos}\mathcal{L}_{pos} + \lambda_{flat}\mathcal{L}_{flat}.
\end{equation}

\noindent{\textbf{Texture Optimization.}}\label{sec:tex opt} After obtaining the geometry with the coarse texture, we continue to learn high-detailed dense texture. Compared to representing geometry, 
generating an \textbf{8K} texture map with pore-level details requires more Gaussians 
to learn from high-resolution images. Different from vanilla 3DGS, which adopts an adaptive densification process that results in a messy topology, we propose a novel densification method, \ie, \textbf{UV Space Densification}, which allows for topologically densified Gaussians corresponding to the UV space of the texture map.
Specifically, at frame $t$, we first initialize the dense Gaussian Mesh $G^{\prime}_t$  with the Gaussians in $G_t$.
Then, we insert each quadrilateral grid in $G^{\prime}_t$ with ($N \times N$) smaller grids by bi-linear interpolation, with each new Gaussian on each inserted vertex. The attributes and UV coordinates of these new Gaussians are also bi-linear interpolated with Gaussians in $G_t$.
We optimize dense Gaussian Mesh $G^{\prime}_t$ by Eq.~\ref{loss:image} and follow the same practice as Sec.~\ref{sec:Gaussian Mesh} to learn high-frequency texture with sub-micron details.

\subsection{Extracting Geometry and Texture from Gaussians}
\label{sec:GTE}
After completing the geometry and texture optimization, we can extract 4D meshes from $\{G_i\}_{i=0}^{F-1}$ and 8K texture maps from $\{G_i^{\prime}\}_{i=0}^{F-1}$.

\noindent{\textbf{Geometry Extraction.}}
Despite special regularization for the scale of Gaussian, it still has volume, resulting in a slight decrease in geometry reconstructed by directly extracting Gaussians' positions. Therefore, we propose Gaussian Normal Expansion to make surfaces derived from Gaussian surfaces resemble real surfaces more closely.
As illustrated in Fig.~\ref{fig:pipeline}~(c), we offset each Gaussian in the direction of the vertex normal by its projection scale in normal direction to obtain the final meshes $\{\mathbf{S}_i\}_{i=0}^{F-1}$.

\noindent{\textbf{Texture Extraction.}}
To generate 8K texture maps from a dense Gaussian mesh $\{G_i^{\prime}\}_{i=0}^{F-1}$, we map Gaussians' colors to UV space based on their UV coordinates. Since our Gaussian meshes are topologized, we can triangulate them and map learned dense texture colors to texture maps $\{\mathbf{M}_i\}_{i=0}^{F-1}$ by a rasterization-based forward rendering method~\cite{NVDiffrast}.

%% file: Figures/pipeline.tex
\begin{figure*}[t]
    \centering
    \includegraphics[width=1\linewidth]{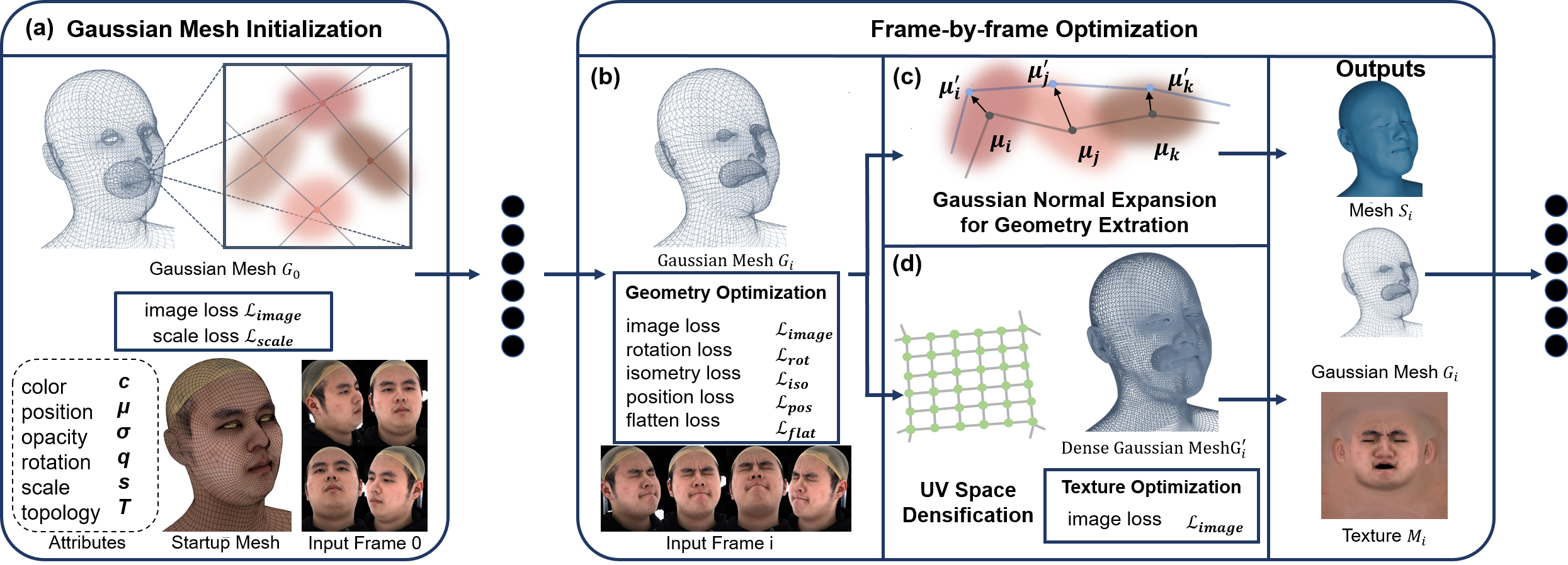}
    \captionof{figure}{Overall pipeline of our framework. \textbf{(a)} We initialize Gaussian attributes and establish topological correspondence with the startup mesh. \textbf{(b)} Take one frame as an example, geometry-related attributes in the Gaussian Mesh of the last frame are optimized by this frame under a set of topology-aware loss items. \textbf{(c)} We align the Gaussian surface with the rendering surface by Gaussian normal expansion to extract more precise meshes. \textbf{(d)} To learn pore-level detailed colors and generate ultra-high resolution texture, we build a dense mesh by densifying Gaussians in UV space.}
    \label{fig:pipeline}
\end{figure*}

%% file: sections/exp.tex
\section{Experiments}
\subsection{Dataset and Implementation Details}
\noindent{\textbf{Data Preparation.}}
We collect a dynamic multi-view head dataset using a Light Stage~\cite{debevec2012light} with 16 calibrated color cameras. In the dataset, images are captured at a resolution of $4096\times3000$ and a rate of 60 fps.
We capture multi-view videos for 10 identities. Each identity should perform an expression sequence and a talking sequence separately, with each expression sequence containing diverse expressions, including extreme and asymmetric ones. Each sequence lasts between 400 to 600 frames and is required to begin from \textbf{a neutral expression. }

\noindent{\textbf{Implementation Details.}}
We implement our method based on PyTorch~\cite{PyTorch} and NVIDIA 3090 GPUs. The mesh topology consists of $n_v=8280$ vertices and $n_f=16494$ faces. 
Since our Light Stage can provide uniform lighting, we use RGB rather than SH in 3DGS to represent view-consistent colors, which is more efficient and easier to optimize.
We use Adam~\cite{Adam} for optimization. The geometry optimization stage includes 1000 iterations at each timestamp, with all images resized to $512\times375$. We mask out the inner mouth using a face parsing model~\cite{face_parsing} to prevent the vertices around the lips from learning incorrect colors.
The texture optimization stage consists of 300 iterations at each timestamp and is learned at the original 4K resolution without preprocessing, with a dense number $N=30$. 
We set $\lambda_{image}=0.2$, $\lambda_{scale}=10$, $\lambda_{rot}=20$, $\lambda_{iso}=20$, $\lambda_{pos}=1e3$, and $\lambda_{flat}=2e-4$.
We employ an MVS method~\cite{MVS} to reconstruct 4D head scans for evaluation and use an automatic iterative closest point (ICP)~\cite{ICP} algorithm to obtain a roughly accurate mesh in the first frame for initialization. 
The texture outside the facial area is obtained by alpha blending with the template textures.
All hyper-parameters remain the \textbf{same} for all experiments. 
Please refer to the Supplementary Material for more details.

\subsection{Face Reconstruction}

\noindent{\textbf{Baseline.}}
We evaluate our method on our collected dataset and compare it to three types of SOTA topology-consistent face reconstruction methods: (1) single-view methods DECA~\cite{DECA},
and HRN~\cite{HRN}; (2) multi-view methods MVFR~\cite{MVFR} and DFNRMVS~\cite{DFNRMVS}; (3) a traditional MVS and ICP pipeline w/wo facial landmark guidance. For single-view methods, we input the front-view image. For multi-view methods, we use all 16 views. Note that, we follow DFNRMVS`s setting that uses a front view and an oblique side view image to produce the best results.

\input{Tables/quantity_cmp}

\noindent{\textbf{Quantitative Comparisons.}}
We evaluate facial reconstruction accuracy using mesh-to-scan distances.
 To minimize inconsistencies among different models, we only compute metrics on the facial region, excluding the ears, back of the head, and neck.
In Tab.~\ref{tbl:quantity_mesh}, our method significantly outperforms other topology-consistent methods in all metrics. Notably, the majority of vertices are located in high-precision ranges, with 52.8\% more vertices within 0.5 mm precision. This improvement is attributed to the introduction of reliable geometry and color priors during Gaussian initialization.

\noindent{\textbf{Qualitative Comparisons.}}
\input{Figures/Face_Reconstruction/quality_cmp_v2}
Fig.~\ref{fig:quality_mesh} shows a visual comparison between meshes reconstructed using different methods. We manually register the faces as the ground truth. Our method can faithfully capture both asymmetric extreme expressions and minor facial changes, outperforming other pre-trained methods. Additionally, we compare our approach with the traditional optimization-based MVS~\cite{MVS} + ICP~\cite{ICP} pipeline, guided by landmarks. Notably, even the advanced landmark detectors produce significant errors under extreme expressions, resulting in semantically incorrect registration, and fine areas are prone to serration and interpenetration. In contrast, our method reconstructs comparable details with correct correspondence.

\subsection{Ultra-High Resolution Texture Generation}
\input{Figures/Texture_Generation/quality_tex}
\input{Figures/Stability/stability}
We qualitatively compare our generated textures with state-of-the-art facial texture estimation models: UnsupTex~\cite{UnsupTex} and HRN~\cite{HRN}. Fig.~\ref{fig:quality_tex} shows the rendering results of different methods, including the 
generated 8K textures. In the comparisons, UnsupTex and HRN are limited to producing low-resolution textures that lack realistic details. In contrast, our method directly generates high-quality 8K textures without up-sampling, faithfully capturing facial wrinkles, hair, and pores, and achieving noticeably superior rendering quality.

\subsection{Temporal Stability Comparisons}
We compare the temporal stability of the generated meshes and textures with other methods. For meshes, We measure geometry consistency by calculating the RMSE between adjacent frames. Fig.~\ref{fig:stability_pic}~(a) shows the curve of each method over an expression sequence. Our method achieves much more temporally stable vertex tracking with extreme expressions, whereas other methods exhibit considerable performance fluctuations. We notice that, in some cases, the inter-frame difference of DECA~\cite{DECA} is lower than ours. This is because DECA fails to capture some extreme expressions or minor facial changes, as shown in Fig.~\ref{fig:quality_mesh}, and therefore tends to keep the meshes unchanged. Instead, our method can faithfully capture facial motions while maintaining temporal consistency.
From the perspective of texture, we measure stability by comparing the PSNR of textures between consecutive frames. As shown in Fig.~\ref{fig:stability_pic}~(b), our method has the highest mean and lowest variance, indicating superior temporal stability.
In conclusion, our method demonstrates better temporal consistency compared to other methods in both geometry and texture.

\subsection{Ablation Studies}
In this section, we assess the impacts of the crucial designs and parameters used in our method. For geometry reconstruction, we ablate all loss items, the number of input views, and the Gaussian normal expansion operation. Regarding texture estimation, we explore the impact of UV densification density on texture quality, along with the influence of $\mathcal{L}_{scale}$.

\noindent{\textbf{Analysis of Loss Items.}}
\input{Figures/Ablation/geo_ablation}
We remove each loss function and keep other settings constant in the ablation study. Fig.~\ref{fig:geo_ablation} shows the qualitative results. 3DGS's~\cite{3DGS} powerful rendering ability allows Gaussians to render similar images with entirely different geometry. We find that it's essential to physically constrain Gaussians' rotation and distance to achieve correct dense correspondence, especially in obstructed areas and areas with dense vertices like lips and nostrils. Although constraining the Gaussian's scale seems to have a limited impact on geometry accuracy (an increase of around 0.1mm), it is crucial for maintaining pore-level texture details and avoiding blurring, as shown in Fig.~\ref{fig:tex_ablation}. The topology prior losses $\mathcal{L}_{pos}$ and $\mathcal{L}_{flat}$ effectively maintain the stability of invisible areas, such as inner sockets, and significantly prevent exaggerated extrapolation or interpenetration. Ablating these mesh smoothing terms may result in lower mesh-to-scan errors, but it compromises the quality of the grid and wiring.

\noindent{\textbf{Analysis of Normal Expansion.}}
Since we constrain the orientation and scale during initialization and optimization, Gaussians should display a small scale in the surface normal direction. As illustrated in Fig.~\ref{fig:geo_ablation}, by using Gaussian normal expansion, we can marginally reduce the overall error.

\noindent{\textbf{Analysis of Input Views.}}
We evaluate the effect of the number of input views on mesh quality, as shown in Fig.~\ref{fig:geo_ablation}.
It's obvious that even with half of the input views, our method can still yield competitive results, demonstrating its applicability to capture systems with fewer cameras. However, when the view number decreases to 4, the reconstruction mesh exhibits significant distortion.

\noindent{\textbf{Analysis of Gaussian Density.}}
\input{Figures/Ablation/tex_ablation}
Fig.~\ref{fig:tex_ablation} depicts the impact of different density levels on texture quality in UV space densification. As the density decreases, the texture becomes blurred and loses detail. Our method can generate textures of any resolution directly. However, higher resolutions require denser Gaussians, more memory overhead, and longer optimization time.

\subsection{Application}
\input{Figures/Appication/application}
Driving digital characters by real performance is widely applied in industry workflows. Our method can retarget the captured expression sequence of an actor faithfully to other characters. Besides, relighting is also a widespread application in the CV and CG pipelines.
Fig.~\ref{fig:application} shows the results of retargeting extreme expressions and wrinkled texture maps learned by our methods to another facial model, as well as relighting rendering results.

%% file: Tables/quantity_cmp.tex
\begin{table*}[t]
\caption{Quantitative evaluation for different face reconstruction methods on our prepared dataset. We measure the percentage of vertices within different error levels and calculate the mean error and variance.}
\centering
\resizebox{\columnwidth}{!}{%
\begin{tabular}{ccccccccc}
\toprule
Type                         & Methods & \textless{}0.2mm(\%)$\uparrow$ & \textless{}0.5mm(\%)$\uparrow$ & \textless{}1mm(\%)$\uparrow$ & \textless{}2mm(\%)$\uparrow$ & \textless{}3mm(\%)$\uparrow$ & Mean(mm)$\downarrow$ & Med.(mm)$\downarrow$ \\ \hline
\multirow{2}{*}{Single-view} & DECA~\cite{DECA}    & 2.055                            & 5.136                            & 10.264                         & 20.075                         & 29.046                         & 8.104                & 5.929                \\
                             & HRN~\cite{HRN}     & 5.170                            & 12.786                           & 20.734                         & 44.692                         & 60.263                         & 2.871                & 2.429                \\ \hline
\multirow{3}{*}{Multi-view}  & MVFR~\cite{MVFR} & 4.139                            & 10.130                           & 19.407                         & 34.629                         & 43.661                         & 7.800                & 4.357                \\
                             & DFNRMVS~\cite{DFNRMVS} & 3.447                            & 8.579                            & 17.064                         & 33.479                         & 48.356                         & 3.649                & 3.214                \\
                            
                             & Ours    & \textbf{22.485}                  & \textbf{52.856}                  & \textbf{87.376}                & \textbf{94.379}                & \textbf{97.697}                & \textbf{0.686}       & \textbf{0.471}       \\ \bottomrule 
\end{tabular}%
}

\label{tbl:quantity_mesh}
\end{table*}

%% file: Figures/Face_Reconstruction/quality_cmp_v2.tex
\begin{figure*}[t]
    \centering
    \includegraphics[width=1\linewidth]{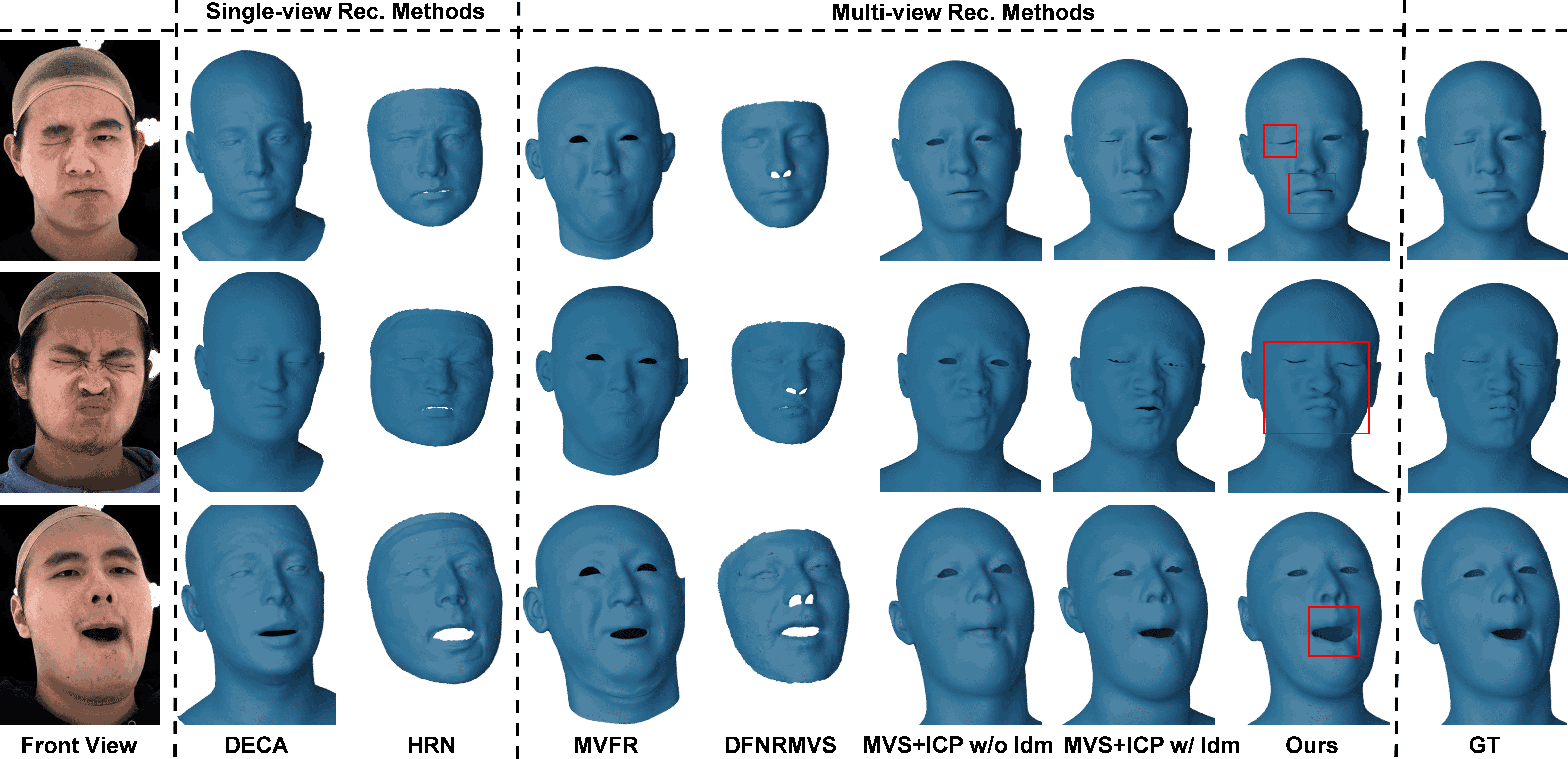}
    \captionof{figure}{Qualitative evaluation of meshes generated by our method and other topology-consistent reconstruction methods. We use artist-manually registered head mesh as the ground truth. We highlight areas that are difficult to reconstruct.
    }
    \label{fig:quality_mesh}
\end{figure*}

%% file: Figures/Texture_Generation/quality_tex.tex
\begin{figure*}[t]
    \centering
    \includegraphics[width=0.9\linewidth]{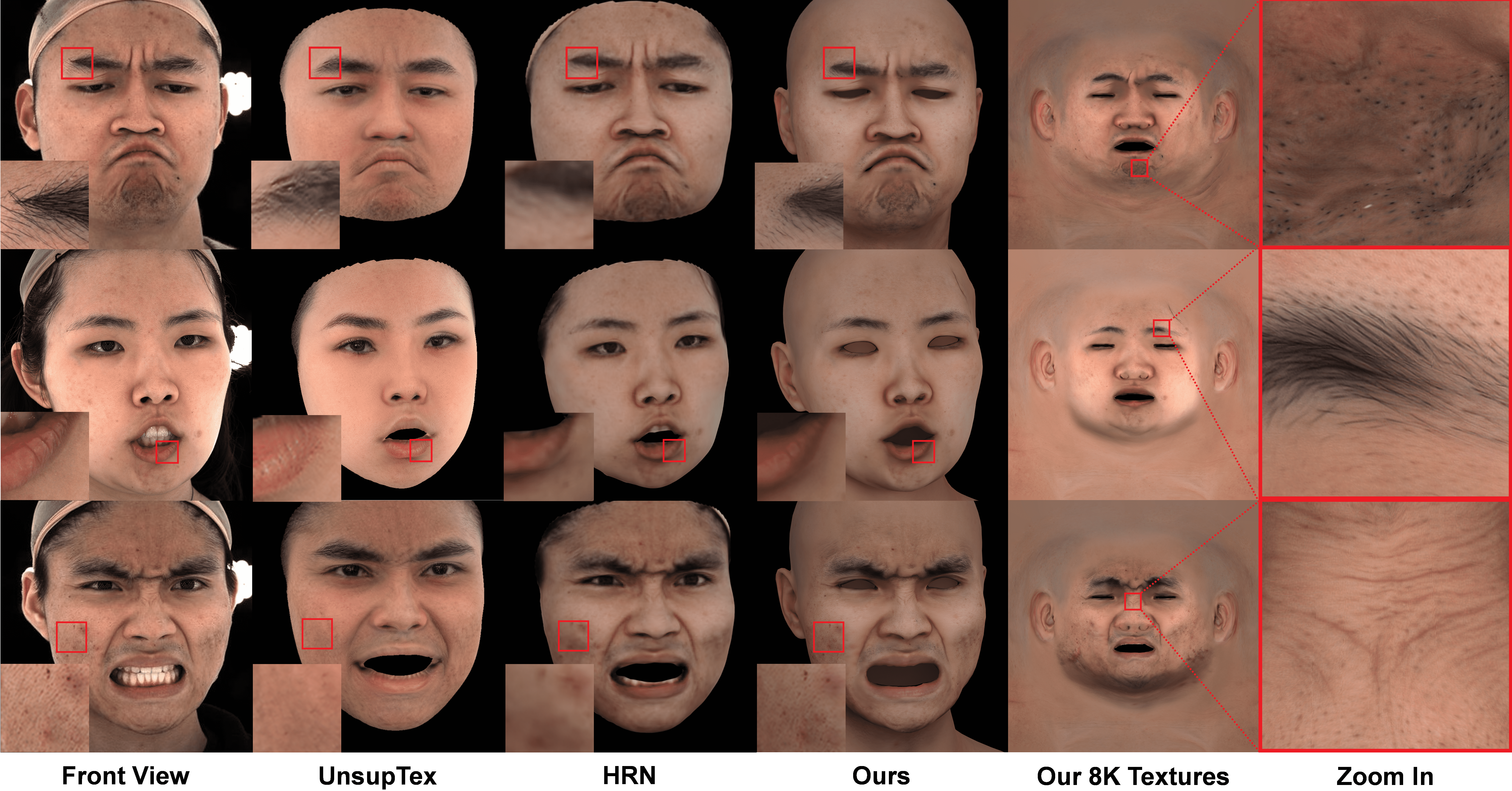}
    \captionof{figure}{Qualitative evaluation of the rendering results between our method, UnsupTex~\cite{UnsupTex} and HRN~\cite{HRN}. 
    The generated 8K textures and pore-level details are demonstrated in columns 5 and 6.}
    \label{fig:quality_tex}
\end{figure*}

%% file: Figures/Stability/stability.tex
\begin{figure*}[t]
    \centering
    \includegraphics[width=0.95\linewidth]{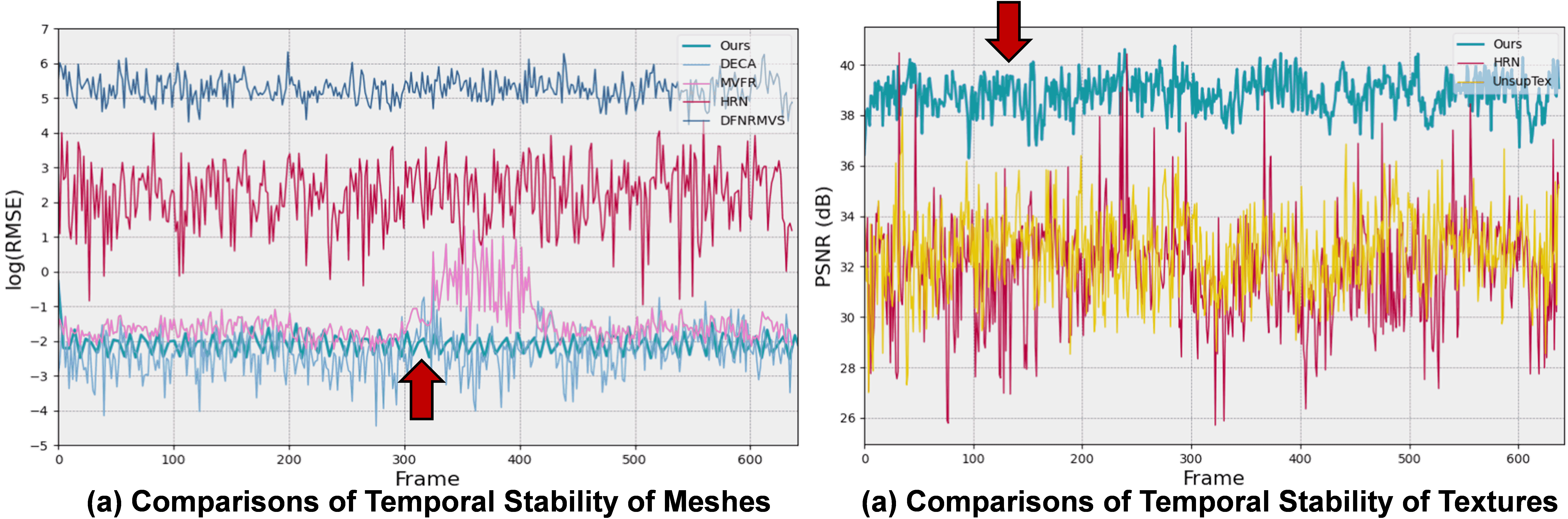}
    \captionof{figure}{Comparisons of temporal stability on a sequence in our dataset, and our method is bolded and indicated by red arrows. \textbf{(a)} The curves of log(RMSE) (lower is better) of several topology-consistent face reconstruction methods. \textbf{(b)} The curves of PSNR (higher is better) are calculated between textures of adjacent frames.}
    \label{fig:stability_pic}
\end{figure*}

%% file: Figures/Ablation/geo_ablation.tex
\begin{figure*}[t]
    \centering \includegraphics[width=1.0\linewidth]{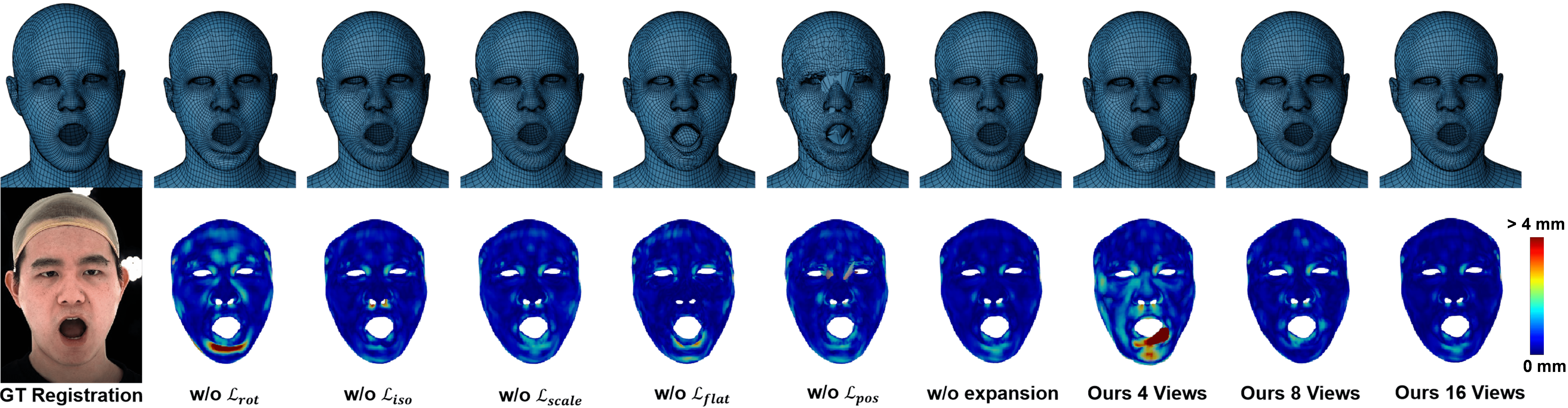}
    \captionof{figure}{Visualization of the reconstructed mesh after ablating loss items, Gaussian normal expansion operation, and view numbers. The second row displays the color-coded point-to-surface
distance between the reconstructed mesh and the scan as a heatmap on the mesh’s surface. \textbf{Please zoom-in for detailed observation.}}
    \label{fig:geo_ablation}
\end{figure*}

%% file: Figures/Ablation/tex_ablation.tex
\begin{figure*}[t]
    \centering
    \includegraphics[width=0.9\linewidth]{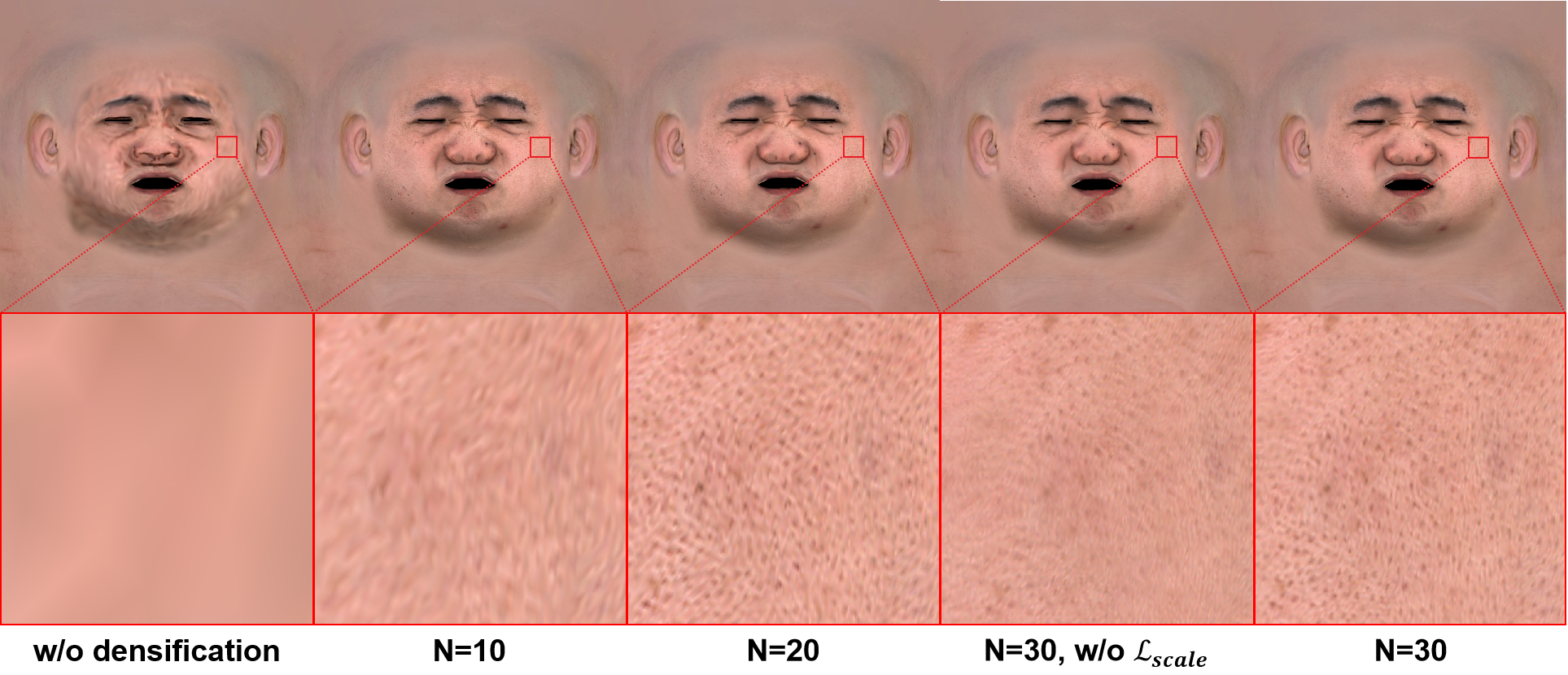}
    \captionof{figure}{Comparisons of texture quality under different settings. 
    }
    \label{fig:tex_ablation}
\end{figure*}

%% file: Figures/Appication/application.tex
\begin{figure*}[t]
    \centering
    \includegraphics[width=0.9\linewidth]{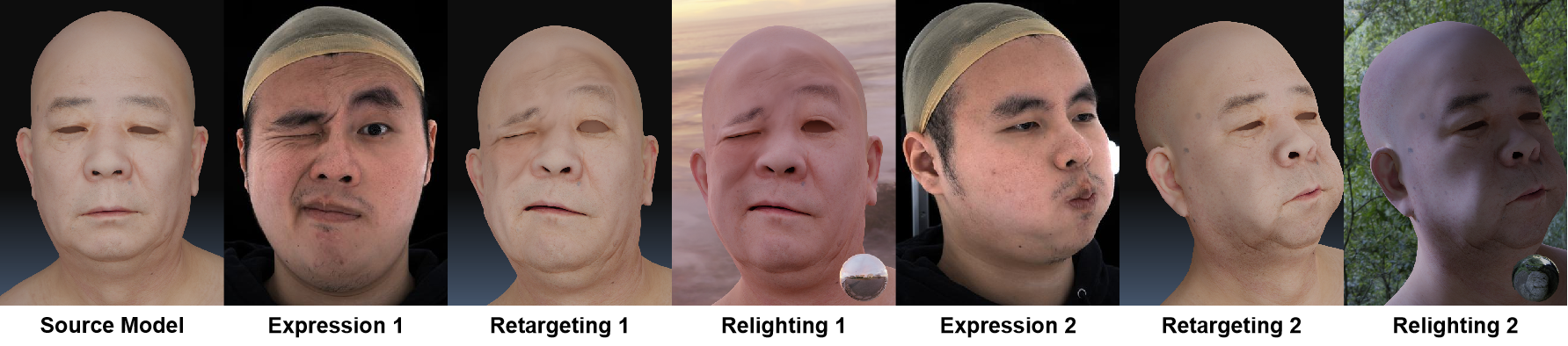}
    \captionof{figure}{We showcase the results of our method applied to retargeting the other face model(columns 3 and 6) and relighting results (columns 4 and 7).}
    \label{fig:application}
\end{figure*}

%% file: sections/conclusion.tex
\section{Conclusion}
In this paper, we propose Topo4D, an efficient framework that can extract temporal topology-consistent meshes and 8K textures from calibrated multi-view videos. Under the regularization of a set of topology-aware geometrical and physical loss items, we achieve topology-preserving Gaussian optimization while faithfully capturing the subject's expressions. By densifying Gaussians in UV space, we learn realistic pore-level details at high resolution and extract high-fidelity 8K texture maps. To sum up, our method provides a brand new way to reconstruct high-fidelity facial meshes and 8K texture maps, opening up new avenues for capturing 4D digital humans in an efficient and low-cost manner.

%% file: supp.tex
\input{sections/supp/ImplDetails}

\input{sections/supp/Exps}

\input{sections/supp/Limitations}
\input{sections/supp/Ethics}

%% file: sections/supp/ImplDetails.tex
\section{Capture System Setup}
\input{Figures/supp/system}
As shown in Fig.~\ref{fig:system}~(a), our \textit{Light Stage} system features 4D data capture capabilities, consisting of 24 time-consistent dynamic cameras capable of capturing multi-view videos at 60FPS with a resolution of $4096\times3000$ pixels. 
Each camera is hardware-controlled with a time error of less than 1 microsecond. 
The 24 cameras are precisely calibrated to obtain accurate intrinsic and extrinsic parameters. 
To minimize perspective errors caused by facial shadows, we employ multiple surrounding light sources.
Subjects are constrained to chairs during performances, maintaining relatively static body movements, which aligns with the facial capture requirements in industrial production processes.

During filming, we capture videos for 10 identities with 2 segments of videos each, lasting approximately 20 seconds for a segment.
In one segment, the subjects spontaneously recite a lengthy passage to simulate natural speech conditions. In the other segment, the subjects randomly change expressions, with many expressions being extreme, involving severe facial distortions and wrinkles, to validate the algorithm's generality and adaptability to extreme scenarios. We show some examples in Fig.~\ref{fig:system}~(b).

\section{Implementation Details}
\subsection{Gaussian Normal Expansion}
The Gaussian function in 3D space corresponds to an ellipsoid. Consequently, the surface shaped by the Gaussian function differs from the geometric surface formed by the Gaussian mean position, which the Gaussian functions generally encase. We thus shift each Gaussian in the direction of the vertex normal to compensate for the gap caused by Gaussian's scale and finally obtain the final meshes. Specifically, we offset each Gaussian's mean position $\bm{\mu}$ by its projection distance from the surface of the ellipsoid in the vertex normal direction:
\begin{align}
    \bm{n}^{\prime} &= R^{-1}\bm{n} \\
    \bm{\mu}^{\prime} &= \bm{\mu} + \sqrt{\frac{1}{\frac{n_x^{\prime 2}}{s_x^2}+\frac{n_y^{\prime 2}}{s_y^2}+\frac{n_z^{\prime 2}}{s_z^2}} }\bm{n},
\end{align}
where $\bm{n}$ is the corresponding vertex normal of each Gaussian.
\subsection{UV Space Densification}
When performing UV Space Densification, we insert each quadrilateral grid in $G^{\prime}_t$ with ($N \times N$) smaller 
equidistant grids. Specifically, the newly generated Gaussian's position, UV coordinates, and color are obtained by bilinear interpolation sampling the corresponding attributes of the four Gaussians (e.g. $G_{0, 0}^{\prime}$, $G_{0, N-1}^{\prime}$, $G_{N-1, 0}^{\prime}$ and $G_{N, N}^{\prime}$) at the original grid vertices: 
\begin{equation}\label{eq:sampling}\begin{split}
A_{i, j} = \frac{1}{(N-1)\times (N-1)}
\begin{bmatrix}
 N-1-i & i
\end{bmatrix}\\
\begin{bmatrix}
 A_{0,0} & A_{0,N-1}\\
 A_{N-1,0} & A_{N-1,N-1}
\end{bmatrix}
\begin{bmatrix}
 N-1-j\\
 j
\end{bmatrix},
\end{split}
\end{equation} where $A$ can be color, uv coordinate, and position. Most importantly, we can easily establish topological relationships between Gaussians. For example, $G_{i, j}$ is connected to $G_{i-1, j}$, $G_{i+1, j}$, $G_{i, j-1}$ and $G_{i, j+1}$. This process is equivalent to subdividing the grid and inserting more sampling points in UV space.

In implementation, we only perform densification once in the first frame and compute every Gaussian's interpolation weights given by Eq.~\ref{eq:sampling}. In the texture optimization stage of each subsquent frame, we use these weights to calculate the attributes of each Gaussian in $G^{\prime}_t$ for dense Gaussian Mesh initialization.

\subsection{Texture Inverse Mapping}
We perform a rasterization-based inverse mapping operation similar to the forward pass in NVDiffrast~\cite{NVDiffrast} to render texture maps. Specifically, if the Gaussian indices of the visible triangles for a pixel at $(x, y)$ are denoted as $i_{0,1,2}$ and the center of mass is denoted as $w_{0,1,2}$, we can calculate the color $\bm{C}_{x,y}$ of the pixel at $(x, y)$.
That is:
\begin{equation}
    \bm{C}_{x,y}=w_0\bm{c}_{i_0} + w_1\bm{c}_{i_1} + (1-w_0-w_1)\bm{c}_{i_2}.
\end{equation}
\subsection{Learning Strategy}
\input{Tables/supp/Attributes}
As is shown in Tab.~\ref{tbl:attributes}, we adopt different optimizing strategies at different stages in our pipeline. \textbf{1)} When optimizing the initial Gaussian Mesh at the first frame, we only optimize $\bm{q}$ and $\bm{s}$, and keep others fixed. \textbf{2)} During geometry optimization, we optimize these geometry-related attributes ($\bm{\mu}$ and $\bm{q}$) to track Gaussian motions. \textbf{3)} During texture optimization, we initialize dense Gaussian Mesh by sampling each Gaussian's position by Eq. \ref{eq:sampling} and set their opacity to 1.  Since we only need to learn the exact color of the UV space sampling points represented by each Gaussian, we fix the Gaussian scale to be the minimum distance with their one-ring neighbors and only optimize Gaussian's color.

%% file: Figures/supp/system.tex
\begin{figure*}[t]
    \centering
    \includegraphics[width=1\linewidth]{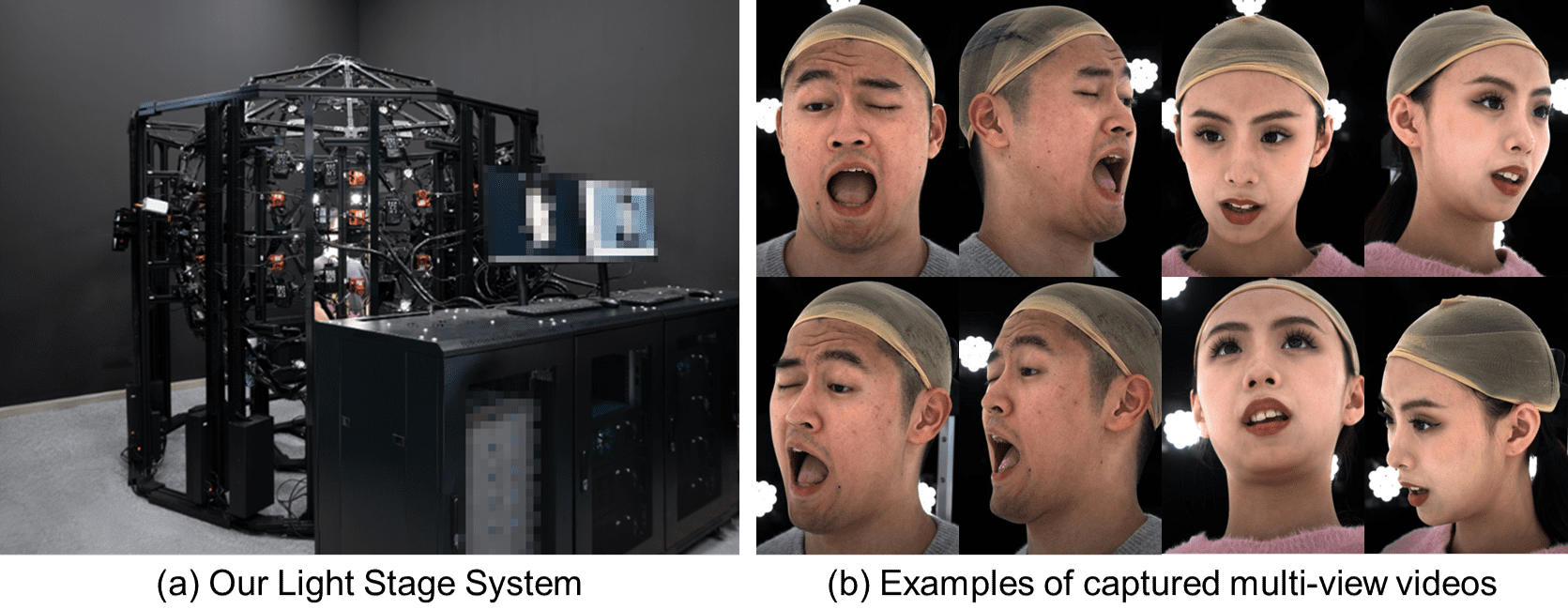}
    \captionof{figure}{We show (a) the diagram of our Light Stage capture system, and (b) some Raw data for different identities.}
    \label{fig:system}
\end{figure*}

%% file: Tables/supp/Attributes.tex
\begin{table}[tb]
\centering
\caption{The initialization settings and learning strategies for each Gaussian attribute.}
\label{tbl:attributes}
\resizebox{\columnwidth}{!}{%
\begin{tabular}{|c|c|c|c|c|c|}
\hline
\textbf{Symbol} & \textbf{Initialization}        & \textbf{First Frame Optimization} & \textbf{Geometry Optimization} & \textbf{Texture Optimization} & \textbf{Learning Rate} \\ \hline
$\bm{\mu}$           & Vertex Coordinate              & Fixed                             & Learnable                      & Fixed                         & 0.000016                \\ \hline
$\bm{q}$             & Vertex Normal                  & Learnable                         & Learnable                      & Fixed                         & 0.001                   \\ \hline
$s$             & Half of Min Neighbour Distance & Learnable                         & Fixed                          & Fixed                         & 0.001                   \\ \hline
$\sigma$        & 1                              & Fixed                             & Fixed                          & Fixed                         & 0.0                     \\ \hline
$\bm{c}$             & Texture Color                  & Fixed                             & Fixed                          & Learnable                     & 0.0025                  \\ \hline
\end{tabular}%
}
\end{table}

%% file: sections/supp/Exps.tex
\section{Additional Experiments}
In this section, we conduct more comparisons with current SOTAs toward geometries and textures. 

\subsection{Additional Geometry Comparisons}
\input{Figures/supp/quality_mesh}
Fig.~\ref{fig:quality_mesh_sup} shows additional qualitative comparisons of our method with current SOTAs, \ie, DECA~\cite{DECA},
HRN~\cite{HRN}, MVFR~\cite{MVFR}, DFNRMVS~\cite{DFNRMVS} and traditional multi-view stereo~\cite{MVS} with iterative closest point~\cite{ICP} (ICP) pipeline. The results indicate that our method outperforms superiorly other approaches. Besides, our method can achieve competitive results with manual registrations while avoiding interpenetration that occurs in the automated ICP method.

\subsection{Additional Texture Comparison}
\input{Figures/supp/quality_tex}
Fig.~\ref{fig:quality_tex_sup} shows additional comparisons of rendering results of textures generated by our method,  UnsupTex~\cite{UnsupTex}, and HRN~\cite{HRN}. These pre-trained model-based methods are unable to handle high-resolution data, and can only generate textures of lower resolution, losing details in rendering results. Benefiting from \textbf{UV Space Densification}, each Gaussian that represents sampling points in UV space can accurately learn pixel-level realistic details from high-resolution inputs.
It's worth mentioning since it is difficult to obtain the same lighting conditions as the capture conditions in the rendering software, there is a certain difference between the rendering results and the captured images. Nonetheless, we also achieve identity consistency and the same details in textures as the images.
Moreover, it can be observed that the generated texture maps maintain realistic wrinkles and pore-level details.

\subsection{Qualitative Comparisons with Current SOTAs}
We next supply comparisons of our method with TEMPEH~\cite{TEMPEH} and ReFA~\cite{REFA}, two state-of-the-art multi-view reconstruction methods. 
Due to different method settings, the mismatch in data structures, or the inaccessibility of codes, it is particularly difficult to compare our method with them. Therefore, we compare them qualitatively in some reasonable settings to show the competitiveness and extensibility of our method, which will be described detailed in the comparisons.

\subsubsection{Comparisons with TEMPEH.}
\input{Figures/supp/tempeh_cmp}

The recent TEMPEH~\cite{TEMPEH}, trained with a large amount of 4D head data, can directly regress dynamic topologic head models from multi-view videos, which shares a similar setting as our work. However, 2 major differences exist between TEMPEH and our Topo4D: \textbf{1)} TEMPEH is trained on its proposed dataset FaMoS, thus it is only applicable to the specific capture system. Hence, new data are required to train the model to utilize TEMPEH in new capture systems.
Conversely, our Topo4D is suitable for more systems with calibrated cameras including FaMoS. 
\textbf{2)} TEMPEH can only generate meshes while our Topo4D can generate both meshes and high-quality textures.

For fair comparisons, we compare TEMPEH and our Topo4D both on our dataset and TEMPEH's FaMoS dataset. Note that, due to the lack of a large amount of registered data in our dataset as supervision, we are unable to train TEMPEH by ourselves. Therefore, we use its publicly released pre-trained models for face reconstruction.
As is shown in Fig.~\ref{fig:tempeh}, \textbf{1)} on our dataset, our method faithfully reconstructs facial geometry and generates 8K textures that can produce realistic rendering results. However, TEMPEH fails to reconstruct meaningful heads, revealing that the pre-trained TEMPEH cannot be directly applied to different capture systems. 
\textbf{2)} FaMoS consists of 16 gray-scale and 8 color images in each frame, including 2 posterolateral color images. In the experiments, we only use 6 color images containing the front face as the input of our method, while showing TEMPEH's best results with its publicly released pre-trained model on 16 gray-scale views. Even though, we achieve competitive geometric results with TEMPEH.
Especially under some extreme conditions in the 4th and 6th columns in Fig.~\ref{fig:tempeh}, we can even better restore asymmetrical eyebrow and pouting expressions. Additionally, our method can generate high-fidelity textures while TEMPEH cannot. 
\textbf{Overall}, our method is more general across capture systems than TEMPEH~\cite{TEMPEH}, and we can generate high-quality texture maps directly. Moreover, our method does not require a large amount of registered data for training.

\subsubsection{Comparisons with ReFA.}
\input{Figures/supp/refa_cmp}
\input{Figures/supp/refa_tex_cmp}
Similar to our Topo4D, ReFA~\cite{REFA} can generate face meshes from multi-view images, while it can generate 4K textures via super-resolution modules. However, since ReFA does not release its codes, pre-trained models, and test datasets, we borrow some results from its paper and website for qualitative comparisons with our Topo4D toward the quality of meshes and texture maps. 
In the geometric comparisons, we select some expressions similar to ReFA's results for better evaluation of extreme expressions.
As shown in Fig.~\ref{fig:refa_cmp}, our method can more faithfully reconstruct the detailed geometric meshes than ReFA, especially in the eyes regions, \eg, 4th and 6th columns. 
As the comparisons on textures shown in Fig.~\ref{fig:refa_tex_cmp}, our Topo4D can faithfully generate 8K high-resolution dynamic textures, including pore-level details and individually discernible strands of hair. Instead, ReFA relies on a super-resolution module to obtain 4K textures, but it fails to faithfully restore the original details of the face and leads to artifacts in the texture.
Overall, our Topo4D achieves better results than ReFA in both mesh fidelity and texture quality.

\subsection{Performance on Multiface Dataset}
\input{Figures/supp/quality_mesh_multiface}
\input{Figures/supp/cmp_tex_multiface}
\input{Figures/rebuttal/multiface2}
To show our method's robustness against different capture systems and extreme expressions, we test our method on Multiface Dataset~\cite{wuu2022multiface} and compare it with other methods, as is shown in Fig.~\ref{fig:quality_mesh_multiface} and Fig.~\ref{fig:quality_tex_multiface}.

We additionally compare our method with two landmark-based optimization methods: \textbf{(1)} Track $+$ Wrap4D~\cite{Wrap4D} guided by a commercial landmark detector, which is widely used in modern CG pipelines. \textbf{(2)}  
Smith et al.~\cite{smith2020constraining}, the tracking pipeline used in Multiface. As is shown in Fig.~\ref{img:multiface2}, current landmark detection methods generally have notable errors under extreme expressions, resulting in wrong correspondence during tracking. Despite using a personalized detector, Smith et al. struggles to track dense areas like eyelids and lips. Due to significant bias in eyelid and lip keypoints, Warp4D produces serrated eyelids, interpenetrated mouth corners, and lips that registered to teeth. In contrast, our method is robust against extreme expressions and outperforms others while keeping stable dense correspondence without using landmarks.

\subsection{Efficiency Comparisons with Optimization-based Methods}
\input{Tables/efficiency_cmp}
As is shown in Tab.~\ref{tbl:efficiency_cmp}, We compare our method with other optimization-based methods from a perspective of computational cost. It is noteworthy that most optimization-based algorithms are not open-sourced, thus we directly use the time cost claimed in their papers. The traditional pipeline (Metashape~\cite{Metashape} $+$ Wrap4D~\cite{Wrap4D}) takes more than 5 minutes to reconstruct a mesh with texture, and requires 
significant more time for manual tweaking. 
The optical-flow based method Fyffe et al.~\cite{fyffe2017multi} reconstructs meshes without MVS but is still more than 20 minutes slower than us when reconstructing a coarse mesh due to its time-consuming volumetric Laplacian solve. Our geometry tracking and 8K texture learning stage both takes only 30 seconds, which is fully automatic and not require any additional supervision such as scans, landmarks, optical flow, etc.

\subsection{Gaussian Mesh Rendering Results}
\input{Figures/supp/GSRender}
Fig.~\ref{fig:gsrender}~(a) shows an example of the Gaussian rendering results of \textit{Gaussian Mesh} and \textit{Dense Gaussian Mesh} under the extreme expression. Our Gaussian Mesh contains only 8280 points for geometry optimization and is learned on $512\times375$ images. Although using such a small number of Gaussian points may cause some degree of blurring in the rendering results, it is sufficient to represent facial geometry and is photometric enough for tracking. Our Dense Gaussian Mesh contains around 5 million points and is learned on raw $4000\times3000$ images. A Gaussian point corresponds to about 5 pixels in the facial UV region, which is nearly enough for 8K texture. It can be observed from the rendering result that such a level of Gaussian density is sufficient to capture pore-level details.

%% file: Figures/supp/quality_mesh.tex
\begin{figure*}[t]
    \centering
    \includegraphics[width=1\linewidth]{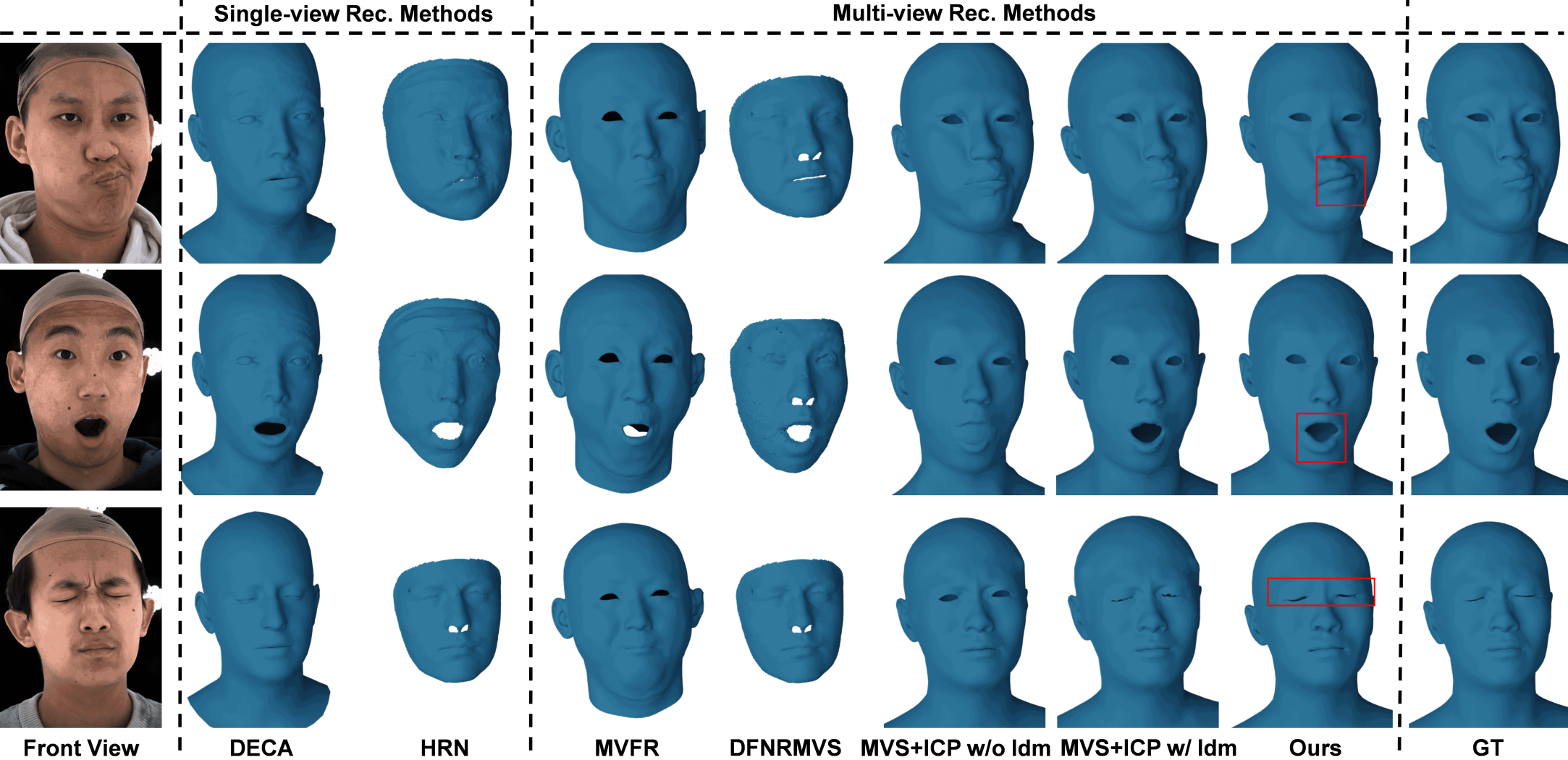}
     \captionof{figure}{Qualitative evaluation of meshes generated by our method and other topology-consistent reconstruction methods. We use artist-manually registered head mesh as the ground truth. We highlight areas that are difficult to reconstruct.
    }
    \label{fig:quality_mesh_sup}
\end{figure*}

%% file: Figures/supp/quality_tex.tex
\begin{figure*}[t]
    \centering
    \includegraphics[width=1\linewidth]{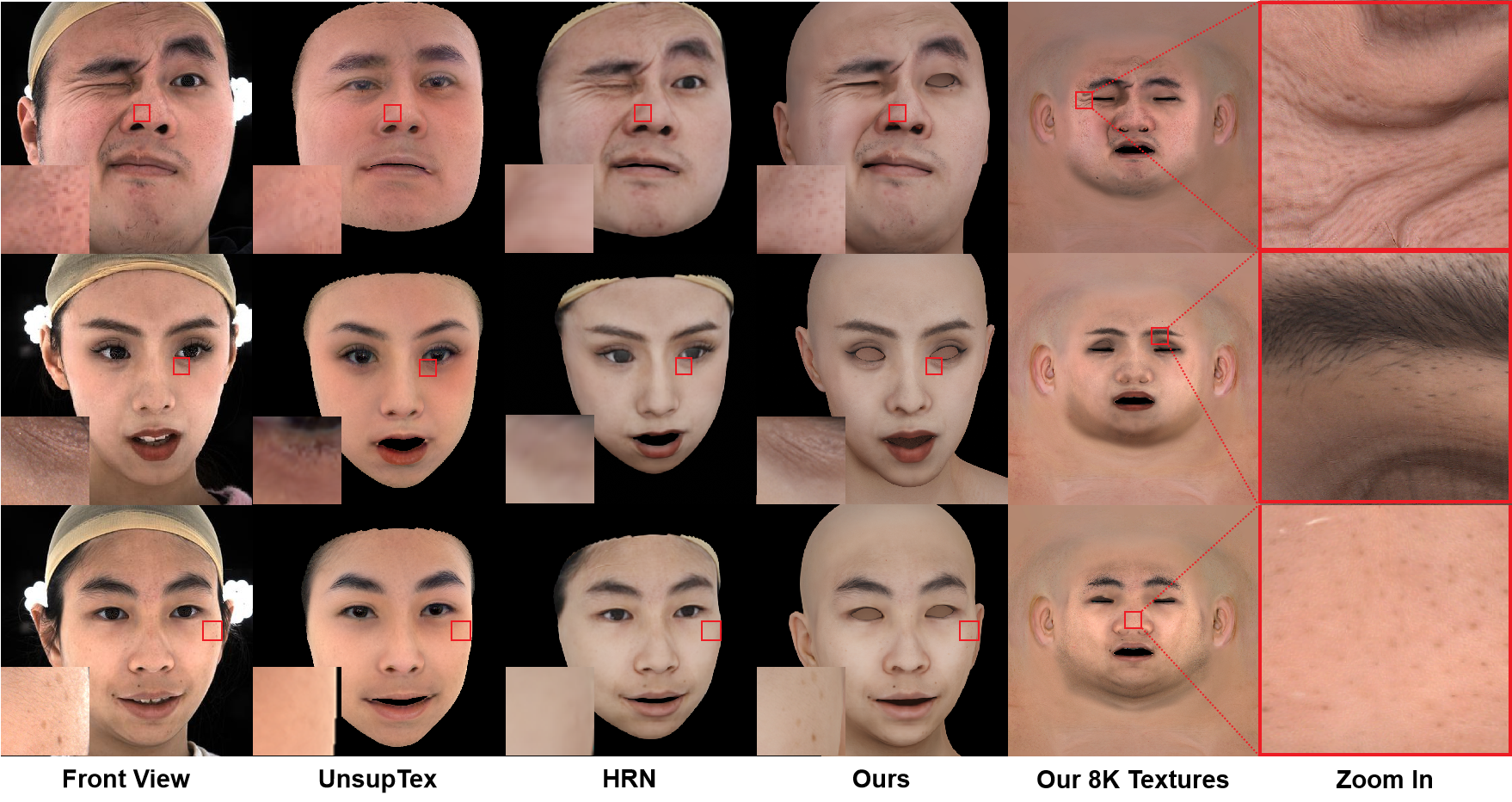}
    \captionof{figure}{Additional qualitative evaluation of the rendering results between our method, UnsupTex~\cite{UnsupTex} and HRN~\cite{HRN}. We also provide the zoom-in renderings for better observation. Our generated 8K textures and pore-level zoom-in details are demonstrated in columns 5 and 6. 
    }
    \label{fig:quality_tex_sup}
\end{figure*}

%% file: Figures/supp/tempeh_cmp.tex
\begin{figure*}[t]
    \centering
    \includegraphics[width=1\linewidth]{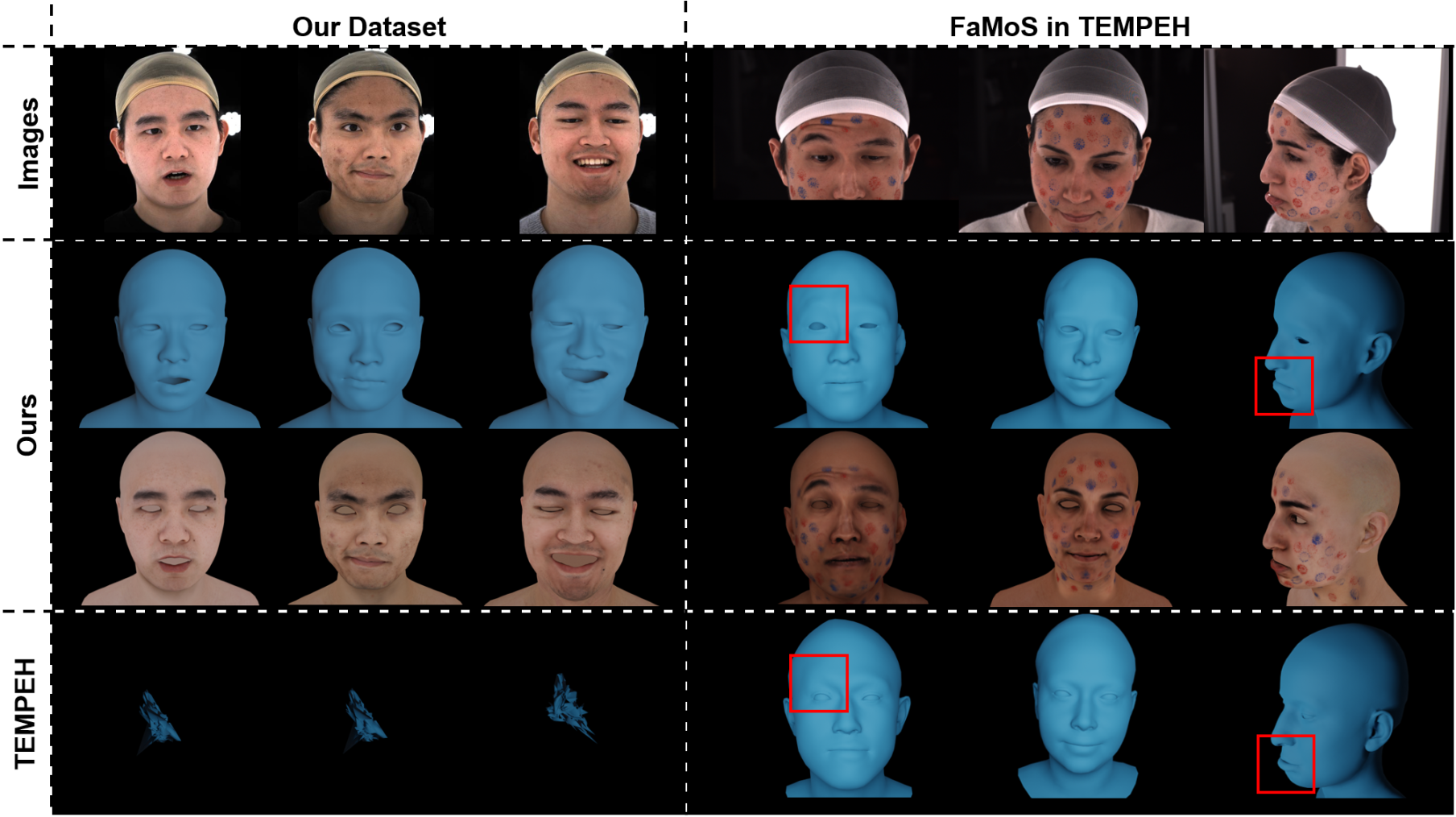}
    \captionof{figure}{Qualitative comparisons with TEMPEH~\cite{TEMPEH} on our dataset and FaMoS. We realize high-quality face reconstruction with textures in both our dataset and FaMoS. TEMPEH achieves competitive geometry results as ours, but it fails to reconstruct facial models in our dataset. Moreover, TEMPEH cannot generate texture maps.}
   
    \label{fig:tempeh}
\end{figure*}

%% file: Figures/supp/refa_cmp.tex
\begin{figure*}[t]
    \centering
    \includegraphics[width=1\linewidth]{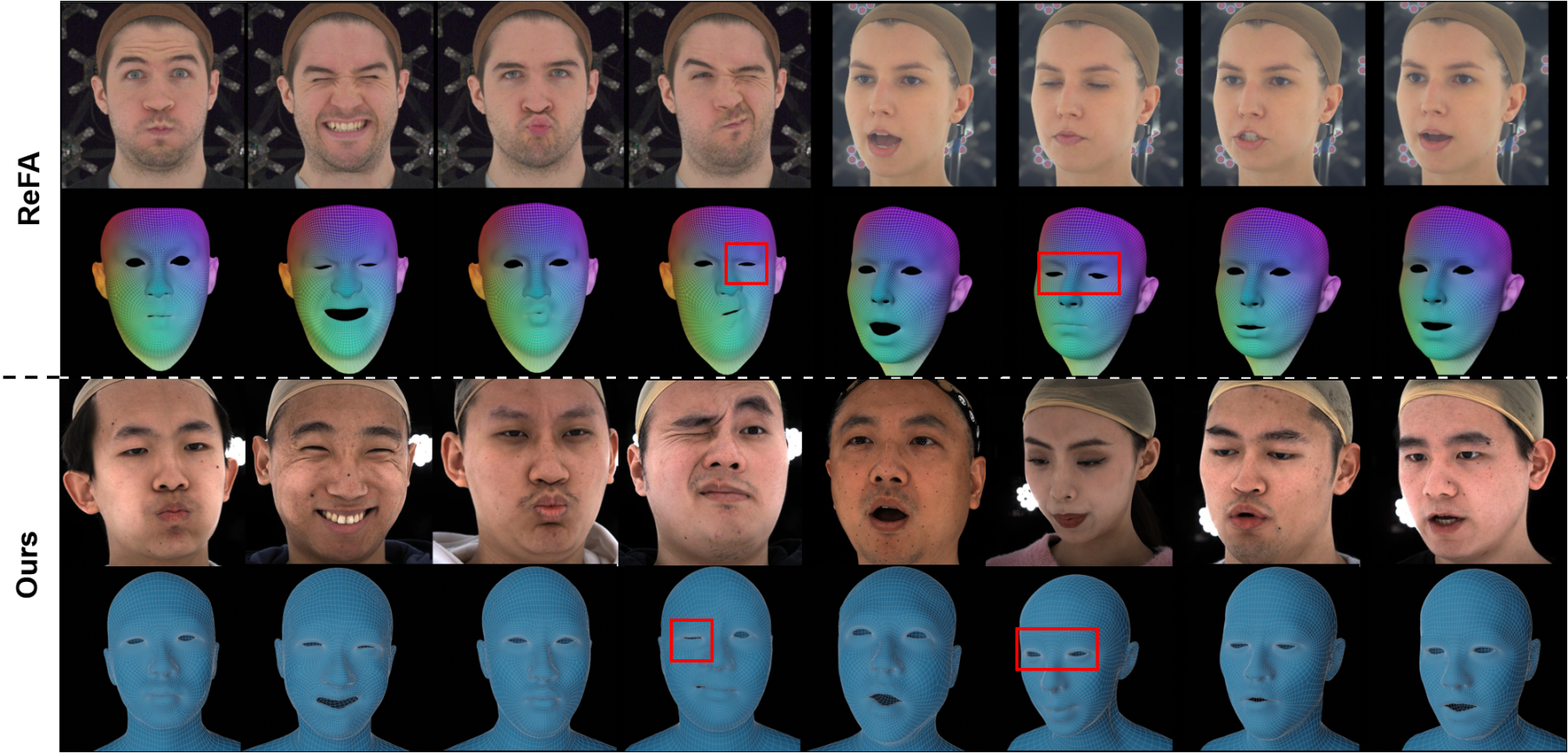}
    \captionof{figure}{Qualitative comparison of geometries with ReFA~\cite{REFA}. We show examples on our dataset similar to those expressions in ReFA's paper. Please \textbf{zoom-in} for detailed observation.}
   
    \label{fig:refa_cmp}
\end{figure*}

%% file: Figures/supp/refa_tex_cmp.tex
\begin{figure*}[t]
    \centering
    \includegraphics[width=1\linewidth]{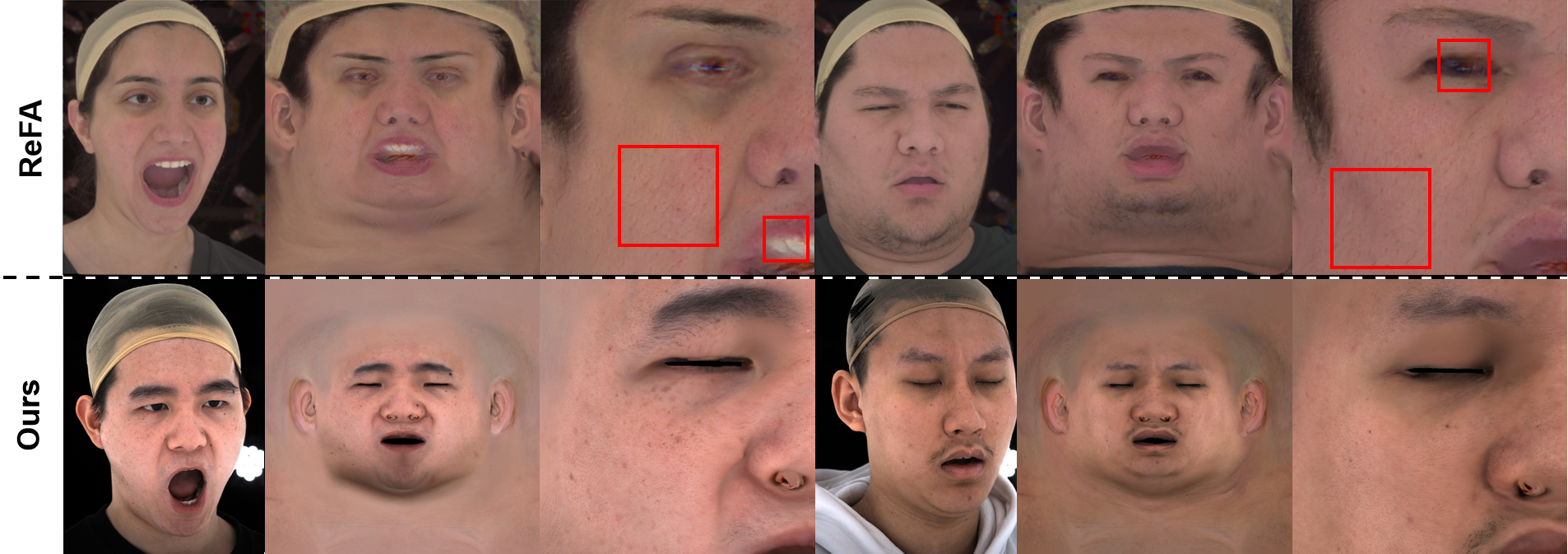}
    \captionof{figure}{Qualitative comparison of textures with ReFA~\cite{REFA}. We show examples of textures generated on our dataset. Please \textbf{zoom-in} for detailed observation.}
   
    \label{fig:refa_tex_cmp}
\end{figure*}

%% file: Figures/supp/quality_mesh_multiface.tex
\begin{figure*}[t]
    \centering
    \includegraphics[width=1\linewidth]{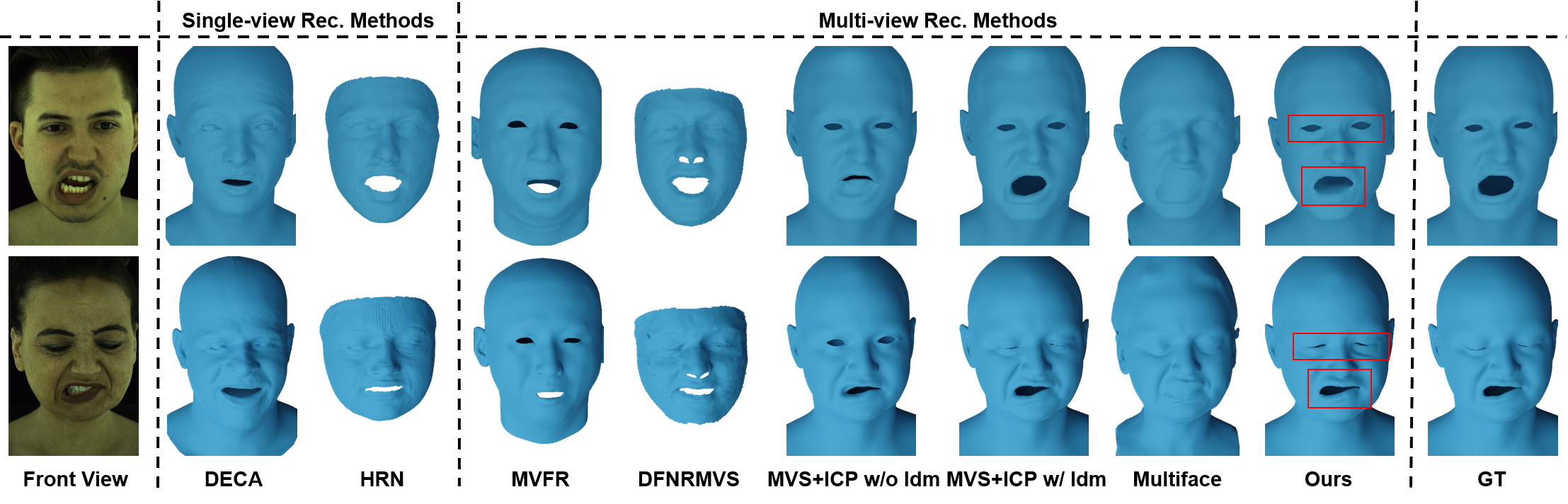}
     \captionof{figure}{Qualitative evaluation of meshes generated by our method and other topology-consistent reconstruction methods on Multiface~\cite{wuu2022multiface} dataset. We use artist-manually registered head mesh as the ground truth. We frame areas that are difficult to reconstruct.
    }
    \label{fig:quality_mesh_multiface}
\end{figure*}

%% file: Figures/supp/cmp_tex_multiface.tex
\begin{figure*}[t]
    \centering
    \includegraphics[width=1\linewidth]{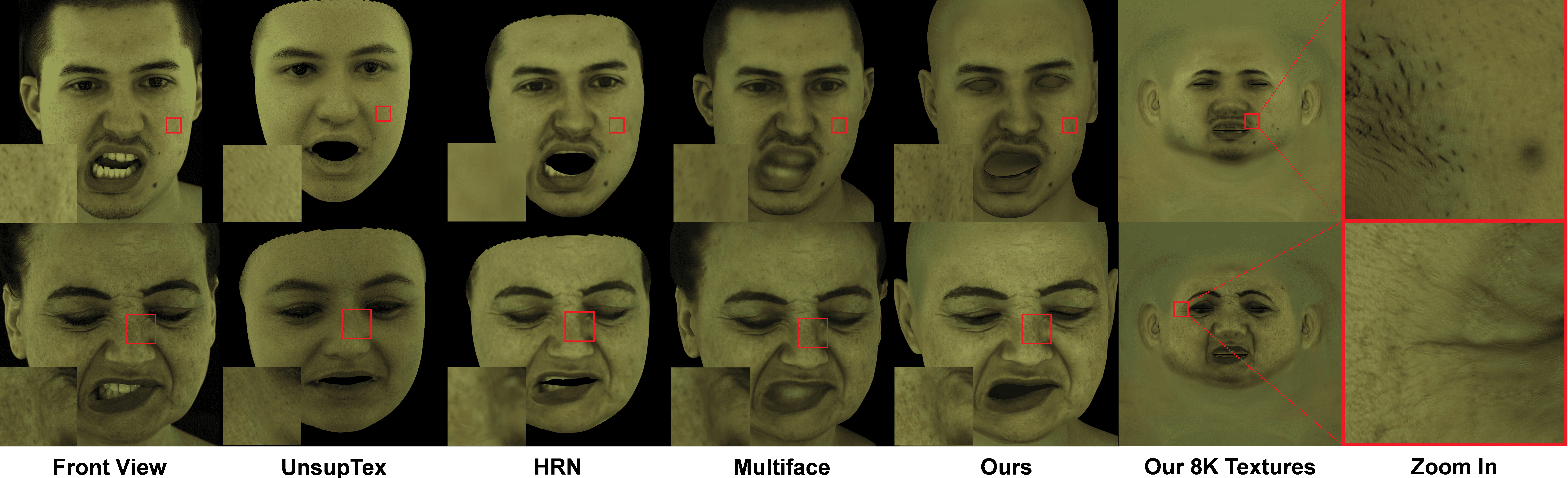}
     \captionof{figure}{Qualitative evaluation of textures generated by our method and other topology-consistent reconstruction methods on Multiface~\cite{wuu2022multiface} dataset. Noting that the images in Multiface dataset are 2K resolution, our method still generates 8K textures.
    }
    \label{fig:quality_tex_multiface}
\end{figure*}

%% file: Figures/rebuttal/multiface2.tex
\begin{figure}[t]
	\centering
	\includegraphics[width=\linewidth]{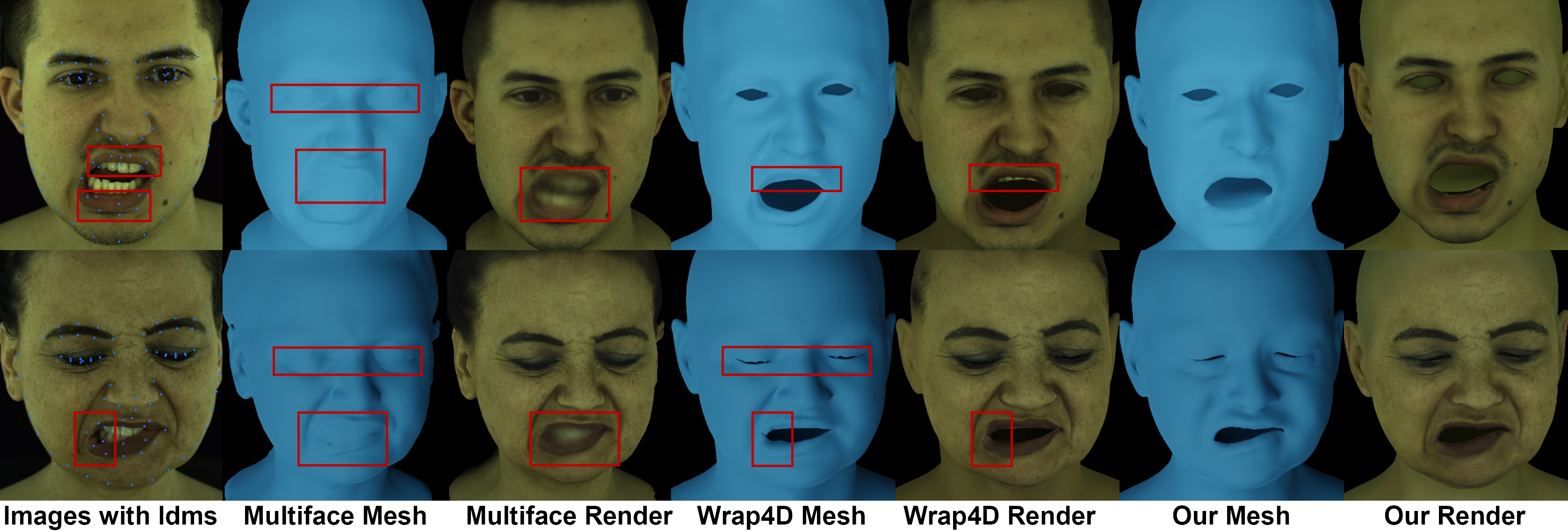}
 \caption{Comparisons on Multiface Dataset. Challenging areas are highlighted in red boxes. Please \textbf{zoom-in}.}
	\label{img:multiface2}
\end{figure}

%% file: Tables/efficiency_cmp.tex
\begin{table}[!htbp]
\caption{Time required per frame of different optimization-based reconstruction methods.}
\centering
\resizebox{1\textwidth}{!}{%
\begin{tabular}{c|c|c|c|c}
\toprule
Method                           & Mesh    & Texture & MVS & Manual\\ 
\midrule
MVS + ICP (Metashape~\cite{Metashape}+Wrap4D~\cite{Wrap4D})           & $\approx$4min & $\approx$80s & $\checkmark$  & $\checkmark$ \\
Fyffe et al.~\cite{fyffe2017multi} & $\approx$25min & -   &  $\times$  &  $\times$  \\
Ours        & $\approx$30s    & $\approx$30s  & $\times$  &  $\times$ \\ \bottomrule
\end{tabular}%
}
\label{tbl:efficiency_cmp}
\end{table}

%% file: Figures/supp/GSRender.tex
\begin{figure*}[t]
    \centering
    \includegraphics[width=1\linewidth]{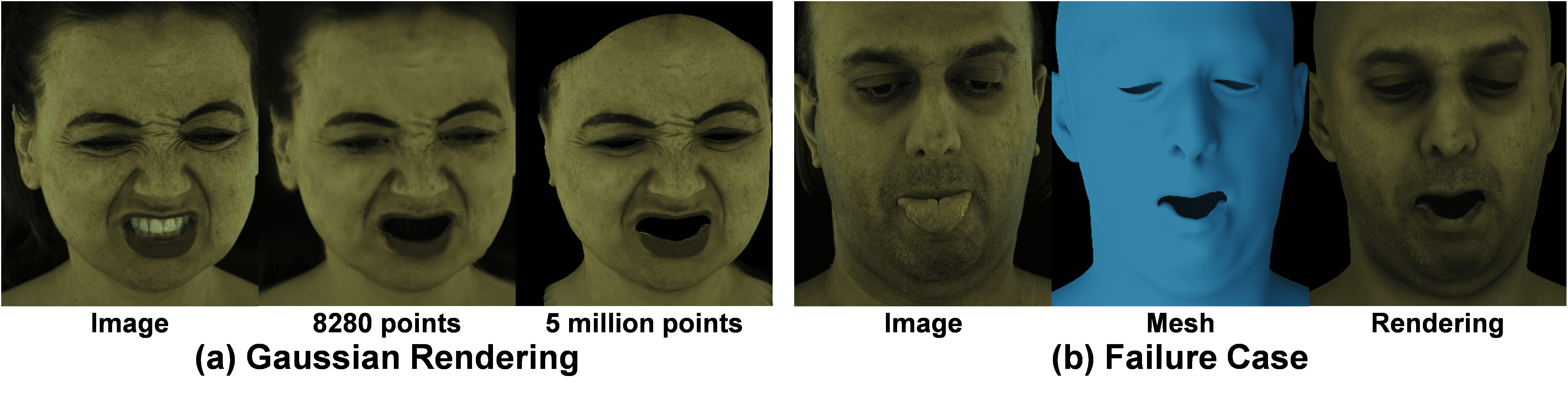}
     \captionof{figure}{(a) Rendering examples of Gaussian Mesh and Dense Gaussian Mesh. (b) An example of tracking failure caused by severe occlusion.
    }
    \label{fig:gsrender}
\end{figure*}

%% file: sections/supp/Limitations.tex
\section{Limitation Discussion}
Topo4D is designed to achieve 4D facial registration without MVS reconstruction and artists' manual intervention. Although it can efficiently and automatically reconstruct 4D facial meshes with pore-level texture details, it still has some limitations. First, our method fails when heavy overlapping occurs due to the lose of tracking, such as sticking out the tongue as is shown in Fig.~\ref{fig:gsrender}~(b), which is commonly solved by artists’ manual operations~\cite{ICTfacekit}. Second, our method inevitably trade-offs between topology quality and surface accuracy, leading to smooth results. We will model detailed geometry by reconstructing displacement maps in future work. Last, due to limited camera angles and the absence of information on polarized light, our method primarily focuses on facial reconstruction and is limited to texture capture only. We aim to extend our method to PBR assets reconstruction in the future.

%% file: sections/supp/Ethics.tex
\section{Ethics Discussion}
In this work, all subjects have signed agreements authorizing us to use the collected data for scientific research on 4D facial reconstruction. We will make every effort to safeguard the original data from disclosure.
Our method relies heavily on visual capture systems similar to the Light Stage for data collection, which can mitigate the risk of misuse to some extent. 
We are committed to privacy protection, preventing the misuse of 4D face reconstruction for criminal purposes.

%% file: main_arxiv.bbl
\begin{thebibliography}{10}
\providecommand{\url}[1]{\texttt{#1}}
\providecommand{\urlprefix}{URL }
\providecommand{\doi}[1]{https://doi.org/#1}

\bibitem{Metashape}
Agisoft metashape: Agisoft metashape. \url{https://www.agisoft.com/}

\bibitem{Wrap4D}
Wrap4d - faceform. \url{https://faceform.com/wrap4d/}

\bibitem{aldrian2012inverse}
Aldrian, O., Smith, W.A.: Inverse rendering of faces with a 3d morphable model.
  TPAMI pp. 1080--1093 (2012)

\bibitem{DFNRMVS}
Bai, Z., Cui, Z., Rahim, J.A., Liu, X., Tan, P.: Deep facial non-rigid
  multi-view stereo. In: CVPR. pp. 5850--5860 (2020)

\bibitem{hifi3dface2021tencentailab}
Bao, L., Lin, X., Chen, Y., Zhang, H., Wang, S., Zhe, X., Kang, D., Huang, H.,
  Jiang, X., Wang, J., Yu, D., Zhang, Z.: High-fidelity 3d digital human head
  creation from rgb-d selfies. TOG  (2021)

\bibitem{barron2021mip}
Barron, J.T., Mildenhall, B., Tancik, M., Hedman, P., Martin-Brualla, R.,
  Srinivasan, P.P.: Mip-nerf: A multiscale representation for anti-aliasing
  neural radiance fields. In: CVPR. pp. 5855--5864 (2021)

\bibitem{barron2022mip}
Barron, J.T., Mildenhall, B., Verbin, D., Srinivasan, P.P., Hedman, P.:
  Mip-nerf 360: Unbounded anti-aliased neural radiance fields. In: CVPR. pp.
  5470--5479 (2022)

\bibitem{bas2017fitting}
Bas, A., Smith, W.A., Bolkart, T., Wuhrer, S.: Fitting a 3d morphable model to
  edges: A comparison between hard and soft correspondences. In: Computer
  Vision--ACCV 2016 Workshops: ACCV 2016 International Workshops, Taipei,
  Taiwan, November 20-24, 2016, Revised Selected Papers, Part II 13. pp.
  377--391 (2017)

\bibitem{beeler2011high}
Beeler, T., Hahn, F., Bradley, D., Bickel, B., Beardsley, P.A., Gotsman, C.,
  Sumner, R.W., Gross, M.H.: High-quality passive facial performance capture
  using anchor frames. ACM Trans. Graph. p.~75 (2011)

\bibitem{ICP}
Besl, P.J., McKay, N.D.: Method for registration of 3-d shapes. In: Sensor
  fusion IV: control paradigms and data structures. pp. 586--606 (1992)

\bibitem{HRN}
Biwen, L., Jianqiang, R., Mengyang, F., Miaomiao, C., Xuansong, X.: A
  hierarchical representation network for accurate and detailed face
  reconstruction from in-the-wild images. In: CVPR. pp. 394--403 (2023)

\bibitem{3DMM}
Blanz, V., Vetter, T.: A morphable model for the synthesis of 3d faces. In:
  SIGGRAPH, pp. 187--194 (1999)

\bibitem{TEMPEH}
Bolkart, T., Li, T., Black, M.J.: Instant multi-view head capture through
  learnable registration. In: CVPR. pp. 768--779 (2023)

\bibitem{booth20163d}
Booth, J., Roussos, A., Zafeiriou, S., Ponniah, A., Dunaway, D.: A 3d morphable
  model learnt from 10,000 faces. In: CVPR. pp. 5543--5552 (2016)

\bibitem{bradley2010high}
Bradley, D., Heidrich, W., Popa, T., Sheffer, A.: High resolution passive
  facial performance capture. In: ACM SIGGRAPH. pp. 1--10 (2010)

\bibitem{cao2013facewarehouse}
Cao, C., Weng, Y., Zhou, S., Tong, Y., Zhou, K.: Facewarehouse: A 3d facial
  expression database for visual computing. TVCG pp. 413--425 (2013)

\bibitem{chen2022tensorf}
Chen, A., Xu, Z., Geiger, A., Yu, J., Su, H.: Tensorf: Tensorial radiance
  fields. In: ECCV. pp. 333--350 (2022)

\bibitem{chen2023monogaussianavatar}
Chen, Y., Wang, L., Li, Q., Xiao, H., Zhang, S., Yao, H., Liu, Y.:
  Monogaussianavatar: Monocular gaussian point-based head avatar. arXiv
  preprint arXiv:2312.04558  (2023)

\bibitem{chen2023mobilenerf}
Chen, Z., Funkhouser, T., Hedman, P., Tagliasacchi, A.: Mobilenerf: Exploiting
  the polygon rasterization pipeline for efficient neural field rendering on
  mobile architectures. In: CVPR. pp. 16569--16578 (2023)

\bibitem{debevec2012light}
Debevec, P.: The light stages and their applications to photoreal digital
  actors. SIGGRAPH Asia pp.~1--6 (2012)

\bibitem{deng2018uv}
Deng, J., Cheng, S., Xue, N., Zhou, Y., Zafeiriou, S.: Uv-gan: Adversarial
  facial uv map completion for pose-invariant face recognition. In: CVPR (2018)

\bibitem{DECA}
Feng, Y., Feng, H., Black, M.J., Bolkart, T.: Learning an animatable detailed
  {3D} face model from in-the-wild images. In: TOG (2021)

\bibitem{fridovich2023k}
Fridovich-Keil, S., Meanti, G., Warburg, F.R., Recht, B., Kanazawa, A.:
  K-planes: Explicit radiance fields in space, time, and appearance. In: CVPR.
  pp. 12479--12488 (2023)

\bibitem{fridovich2022plenoxels}
Fridovich-Keil, S., Yu, A., Tancik, M., Chen, Q., Recht, B., Kanazawa, A.:
  Plenoxels: Radiance fields without neural networks. In: CVPR. pp. 5501--5510
  (2022)

\bibitem{fyffe2017multi}
Fyffe, G., Nagano, K., Huynh, L., Saito, S., Busch, J., Jones, A., Li, H.,
  Debevec, P.: Multi-view stereo on consistent face topology. In: Computer
  Graphics Forum. pp. 295--309 (2017)

\bibitem{gao2021dynamic}
Gao, C., Saraf, A., Kopf, J., Huang, J.B.: Dynamic view synthesis from dynamic
  monocular video. In: CVPR. pp. 5712--5721 (2021)

\bibitem{Ganfit}
Gecer, B., Ploumpis, S., Kotsia, I., Zafeiriou, S.: Ganfit: Generative
  adversarial network fitting for high fidelity 3d face reconstruction. In:
  CVPR (2019)

\bibitem{goesele2006multi}
Goesele, M., Curless, B., Seitz, S.M.: Multi-view stereo revisited. In: CVPR.
  pp. 2402--2409 (2006)

\bibitem{3DDFA}
Guo, J., Zhu, X., Yang, Y., Yang, F., Lei, Z., Li, S.Z.: Towards fast, accurate
  and stable 3d dense face alignment. In: ECCV (2020)

\bibitem{ji2021light}
Ji, P., Li, H., Jiang, L., Liu, X.: Light-weight multi-view topology consistent
  facial geometry and reflectance capture. In: CGI. pp. 139--150 (2021)

\bibitem{jiang2023alignerf}
Jiang, Y., Hedman, P., Mildenhall, B., Xu, D., Barron, J.T., Wang, Z., Xue, T.:
  Alignerf: High-fidelity neural radiance fields via alignment-aware training.
  In: CVPR. pp. 46--55 (2023)

\bibitem{3DGS}
Kerbl, B., Kopanas, G., Leimk{\"u}hler, T., Drettakis, G.: 3d gaussian
  splatting for real-time radiance field rendering. TOG  (2023)

\bibitem{Adam}
Kingma, D.P., Ba, J.: Adam: A method for stochastic optimization. arXiv
  preprint arXiv:1412.6980  (2014)

\bibitem{NVDiffrast}
Laine, S., Hellsten, J., Karras, T., Seol, Y., Lehtinen, J., Aila, T.: Modular
  primitives for high-performance differentiable rendering. TOG  (2020)

\bibitem{Lattas20}
Lattas, A., Moschoglou, S., Gecer, B., Ploumpis, S., Triantafyllou, V., Ghosh,
  A., Zafeiriou, S.: {AvatarMe}: Realistically renderable {3D} facial
  reconstruction. In: CVPR (2020)

\bibitem{lattas2023fitme}
Lattas, A., Moschoglou, S., Ploumpis, S., Gecer, B., Deng, J., Zafeiriou, S.:
  Fitme: Deep photorealistic 3d morphable model avatars. In: CVPR. pp.
  8629--8640 (2023)

\bibitem{ICTfacekit}
Li, R., Bladin, K., Zhao, Y., Chinara, C., Ingraham, O., Xiang, P., Ren, X.,
  Prasad, P., Kishore, B., Xing, J., et~al.: Learning formation of
  physically-based face attributes. In: CVPR. pp. 3410--3419 (2020)

\bibitem{FLAME}
Li, T., Bolkart, T., Black, M.J., Li, H., Romero, J.: Learning a model of
  facial shape and expression from 4d scans. TOG pp. 194--1 (2017)

\bibitem{TOFU}
Li, T., Liu, S., Bolkart, T., Liu, J., Li, H., Zhao, Y.: Topologically
  consistent multi-view face inference using volumetric sampling. In: CVPR. pp.
  3824--3834 (2021)

\bibitem{li2022neural}
Li, T., Slavcheva, M., Zollhoefer, M., Green, S., Lassner, C., Kim, C.,
  Schmidt, T., Lovegrove, S., Goesele, M., Newcombe, R., et~al.: Neural 3d
  video synthesis from multi-view video. In: CVPR. pp. 5521--5531 (2022)

\bibitem{li2023animatable}
Li, Z., Zheng, Z., Wang, L., Liu, Y.: Animatable gaussians: Learning
  pose-dependent gaussian maps for high-fidelity human avatar modeling. arXiv
  preprint arXiv:2311.16096  (2023)

\bibitem{liang2023gaufre}
Liang, Y., Khan, N., Li, Z., Nguyen-Phuoc, T., Lanman, D., Tompkin, J., Xiao,
  L.: Gaufre: Gaussian deformation fields for real-time dynamic novel view
  synthesis. arXiv preprint arXiv:2312.11458  (2023)

\bibitem{lin2022efficient}
Lin, H., Peng, S., Xu, Z., Yan, Y., Shuai, Q., Bao, H., Zhou, X.: Efficient
  neural radiance fields for interactive free-viewpoint video. In: SIGGRAPH
  Asia. pp.~1--9 (2022)

\bibitem{face_parsing}
Lin, Y., Shen, J., Wang, Y., Pantic, M.: Roi tanh-polar transformer network for
  face parsing in the wild. Image and Vision Computing p. 104190 (2021)

\bibitem{REFA}
Liu, S., Cai, Y., Chen, H., Zhou, Y., Zhao, Y.: Rapid face asset acquisition
  with recurrent feature alignment. TOG pp. 1--17 (2022)

\bibitem{lombardi2021mixture}
Lombardi, S., Simon, T., Schwartz, G., Zollhoefer, M., Sheikh, Y., Saragih, J.:
  Mixture of volumetric primitives for efficient neural rendering. TOG pp.
  1--13 (2021)

\bibitem{MVS}
Loop, C., Zhang, Z.: Computing rectifying homographies for stereo vision. In:
  CVPR. pp. 125--131 (1999)

\bibitem{DynamicGaussian}
Luiten, J., Kopanas, G., Leibe, B., Ramanan, D.: Dynamic 3d gaussians: Tracking
  by persistent dynamic view synthesis. arXiv preprint arXiv:2308.09713  (2023)

\bibitem{ma2007rapid}
Ma, W.C., Hawkins, T., Peers, P., Chabert, C.F., Weiss, M., Debevec, P.E.,
  et~al.: Rapid acquisition of specular and diffuse normal maps from polarized
  spherical gradient illumination. Rendering Techniques p.~10 (2007)

\bibitem{meng2023neat}
Meng, X., Chen, W., Yang, B.: Neat: Learning neural implicit surfaces with
  arbitrary topologies from multi-view images. In: CVPR. pp. 248--258 (2023)

\bibitem{mildenhall2021nerf}
Mildenhall, B., Srinivasan, P.P., Tancik, M., Barron, J.T., Ramamoorthi, R.,
  Ng, R.: Nerf: Representing scenes as neural radiance fields for view
  synthesis. Communications of the ACM pp. 99--106 (2021)

\bibitem{muller2022instant}
M{\"u}ller, T., Evans, A., Schied, C., Keller, A.: Instant neural graphics
  primitives with a multiresolution hash encoding. TOG pp. 1--15 (2022)

\bibitem{PyTorch}
Paszke, A., Gross, S., Massa, F., Lerer, A., Bradbury, J., Chanan, G., Killeen,
  T., Lin, Z., Gimelshein, N., Antiga, L., et~al.: Pytorch: An imperative
  style, high-performance deep learning library. NeurIPS  \textbf{32} (2019)

\bibitem{ploumpis2020towards}
Ploumpis, S., Ververas, E., O'Sullivan, E., Moschoglou, S., Wang, H., Pears,
  N., Smith, W.A., Gecer, B., Zafeiriou, S.: Towards a complete 3d morphable
  model of the human head. TPAMI pp. 4142--4160 (2020)

\bibitem{GaussianAvatar}
Qian, S., Kirschstein, T., Schoneveld, L., Davoli, D., Giebenhain, S.,
  Nie{\ss}ner, M.: Gaus sianavatars: Photorealistic head avatars with rigged 3d
  gaussians. arXiv preprint arXiv:2312.02069  (2023)

\bibitem{qian20233dgs}
Qian, Z., Wang, S., Mihajlovic, M., Geiger, A., Tang, S.: 3dgs-avatar:
  Animatable avatars via deformable 3d gaussian splatting. arXiv preprint
  arXiv:2312.09228  (2023)

\bibitem{ren2023facial}
Ren, X., Lattas, A., Gecer, B., Deng, J., Ma, C., Yang, X.: Facial geometric
  detail recovery via implicit representation. In: 2023 IEEE 17th International
  Conference on Automatic Face and Gesture Recognition (FG). pp.~1--8. IEEE
  (2023)

\bibitem{UnsupTex}
Slossberg, R., Jubran, I., Kimmel, R.: Unsupervised high-fidelity facial
  texture generation and reconstruction. In: ECCV (2022)

\bibitem{smith2020constraining}
Smith, B., Wu, C., Wen, H., Peluse, P., Sheikh, Y., Hodgins, J.K., Shiratori,
  T.: Constraining dense hand surface tracking with elasticity. ACM
  Transactions on Graphics (ToG)  \textbf{39}(6),  1--14 (2020)

\bibitem{thies2016face2face}
Thies, J., Zollhofer, M., Stamminger, M., Theobalt, C., Nie{\ss}ner, M.:
  Face2face: Real-time face capture and reenactment of rgb videos. In: CVPR.
  pp. 2387--2395 (2016)

\bibitem{wang2021neus}
Wang, P., Liu, L., Liu, Y., Theobalt, C., Komura, T., Wang, W.: Neus: Learning
  neural implicit surfaces by volume rendering for multi-view reconstruction.
  NeurIPS pp. 27171--27183 (2021)

\bibitem{wood20223d}
Wood, E., Baltru{\v{s}}aitis, T., Hewitt, C., Johnson, M., Shen, J.,
  Milosavljevi{\'c}, N., Wilde, D., Garbin, S., Sharp, T., Stojiljkovi{\'c},
  I., et~al.: 3d face reconstruction with dense landmarks. In: ECCV. pp.
  160--177 (2022)

\bibitem{MVFNet}
Wu, F., Bao, L., Chen, Y., Ling, Y., Song, Y., Li, S., Ngan, K.N., Liu, W.:
  Mvf-net: Multi-view 3d face morphable model regression. In: CVPR. pp.
  959--968 (2019)

\bibitem{wu20234d}
Wu, G., Yi, T., Fang, J., Xie, L., Zhang, X., Wei, W., Liu, W., Tian, Q., Wang,
  X.: 4d gaussian splatting for real-time dynamic scene rendering. arXiv
  preprint arXiv:2310.08528  (2023)

\bibitem{wuu2022multiface}
Wuu, C.h., Zheng, N., Ardisson, S., Bali, R., Belko, D., Brockmeyer, E., Evans,
  L., Godisart, T., Ha, H., Huang, X., et~al.: Multiface: A dataset for neural
  face rendering. arXiv preprint arXiv:2207.11243  (2022)

\bibitem{MVFR}
Xiao, Y., Zhu, H., Yang, H., Diao, Z., Lu, X., Cao, X.: Detailed facial
  geometry recovery from multi-view images by learning an implicit function.
  In: AAAI (2022)

\bibitem{xu2023gaussian}
Xu, Y., Chen, B., Li, Z., Zhang, H., Wang, L., Zheng, Z., Liu, Y.: Gaussian
  head avatar: Ultra high-fidelity head avatar via dynamic gaussians. arXiv
  preprint arXiv:2312.03029  (2023)

\bibitem{dialoguenerf}
Yan, Y., Zhou, Z., Wang, Z., Gao, J., Yang, X.: Dialoguenerf: Towards realistic
  avatar face-to-face conversation video generation. arXiv preprint
  arXiv:2203.07931  (2022)

\bibitem{yang2020facescape}
Yang, H., Zhu, H., Wang, Y., Huang, M., Shen, Q., Yang, R., Cao, X.: Facescape:
  a large-scale high quality 3d face dataset and detailed riggable 3d face
  prediction. In: CVPR. pp. 601--610 (2020)

\bibitem{yang2023real}
Yang, Z., Yang, H., Pan, Z., Zhu, X., Zhang, L.: Real-time photorealistic
  dynamic scene representation and rendering with 4d gaussian splatting. arXiv
  preprint arXiv:2310.10642  (2023)

\bibitem{yang2023deformable}
Yang, Z., Gao, X., Zhou, W., Jiao, S., Zhang, Y., Jin, X.: Deformable 3d
  gaussians for high-fidelity monocular dynamic scene reconstruction. arXiv
  preprint arXiv:2309.13101  (2023)

\bibitem{yariv2023bakedsdf}
Yariv, L., Hedman, P., Reiser, C., Verbin, D., Srinivasan, P.P., Szeliski, R.,
  Barron, J.T., Mildenhall, B.: Bakedsdf: Meshing neural sdfs for real-time
  view synthesis. arXiv preprint arXiv:2302.14859  (2023)

\bibitem{zhang2022video}
Zhang, L., Zeng, C., Zhang, Q., Lin, H., Cao, R., Yang, W., Xu, L., Yu, J.:
  Video-driven neural physically-based facial asset for production. TOG pp.
  1--16 (2022)

\bibitem{zielonka2023drivable}
Zielonka, W., Bagautdinov, T., Saito, S., Zollh{\"o}fer, M., Thies, J., Romero,
  J.: Drivable 3d gaussian avatars. arXiv preprint arXiv:2311.08581  (2023)

\end{thebibliography}
